\documentclass[twoside,11pt]{article}

\usepackage{amsfonts}
\usepackage{bm}
\usepackage{url}
\usepackage{tikz}
\usetikzlibrary{calc,arrows.meta}
\usepackage{amsmath}
\usepackage{a4wide}
\usepackage{apxproof}
\usepackage{amsthm}
\usepackage{mdframed}
\usepackage{accents}
\usepackage{ifthen}
\usepackage[most]{tcolorbox}
\usepackage{amssymb}
\usepackage{ulem}
\usepackage{xcolor}
\usepackage{xifthen}

\usepackage[preprint]{jmlr_arxiv}
\usepackage{cancel}

\newtcolorbox[auto counter]{boxedexample}[2][]{
	enhanced,
	colframe=gray!25,
	colback=gray!5,
	coltitle=black,
	fonttitle=\bfseries,
	breakable, % Allow breaking across pages
	sharp corners, % Cleaner corners
	boxrule=0.5mm, % Thickness of the frame
	width=\textwidth, % Box width
	leftrule=1mm, % Left border thickness
	toprule=1mm, % Top border thickness
	bottomrule=1mm, % Bottom border thickness
	title=Example~\thetcbcounter. #1,
	parbox=false,
	before upper={\parindent\parindent},
	label = #2
}

\newcommand{\vect}[1]{\mathbf{#1}}
\newcommand{\opt}[1]{#1^*}
\newcommand{\lab}{t}
\newcommand{\rlab}{T}
\newcommand{\vrlab}{\vect{\rlab}}
\newcommand{\vlab}{\vect{\lab}}
\newcommand{\vlabopt}{\opt{\vlab}}
\newcommand{\pred}{y}
\newcommand{\rpred}{Y}
\newcommand{\vrpred}{\vect{\rpred}}
\newcommand{\vpred}{\vect{\pred}}
\newcommand{\vpredopt}{\opt{\vpred}}
\newcommand{\vectg}{\vect{g}}
\newcommand{\vectf}{\vect{f}}
\newcommand{\vphi}{\boldsymbol{\phi}}
\newcommand{\bvlab}{\bar{\vlab}}
\newcommand{\tvlab}{\underaccent{\bar}{\vlab}}
\newcommand{\bvpred}{\bar{\vpred}}

\newcommand{\veps}{\boldsymbol{\epsilon}}
\newcommand{\variance}{v}
\newcommand{\rvariance}{V}
\newcommand{\expectation}{\mathbb{E}}
\newcommand{\argmin}{\operatornamewithlimits{arg\,min}}
\newcommand{\Rplus}{\mathbb{R}_{\geq 0}}
\newcommand{\labelspace}{\mathcal{Y}}
\newcommand{\loss}{\ell}
\newcommand{\hessian}{\mathcal{L}}
\newcommand{\gradient}{\boldsymbol{\ell}}
\newcommand{\bregman}[1]{\mathcal{D}_{#1}}
\newcommand{\gbregman}[2]{\mathcal{D}_{#1,#2}}
\newcommand{\vectb}{\vect{b}}
\newcommand{\vlambda}{\boldsymbol{\lambda}}

\newcommand{\vones}{\vect{1}}

\newif\ifshowchanges
\newif\ifshowerrata
\showchangesfalse
\showerratafalse

\definecolor{addedcolor}{rgb}{0,0,1}  % blue
\definecolor{erratacolor}{rgb}{0,0.7,0}   % green
\definecolor{deletedcolor}{rgb}{1,0,0} % red

\DeclareRobustCommand{\added}[1]{%
	\ifshowchanges
		\ifmmode
			\begingroup\color{addedcolor}#1\endgroup%
		\else
			{\color{addedcolor}#1}%
		\fi
	\else
		#1%
	\fi
}

\DeclareRobustCommand{\errata}[1]{%
	\ifshowerrata
		\ifmmode
			\begingroup\color{erratacolor}#1\endgroup%
		\else
			{\color{erratacolor}#1}%
		\fi
	\else
		#1%
	\fi
}

\DeclareRobustCommand{\deleted}[1]{%
	\ifshowchanges
		\ifmmode
			\textcolor{deletedcolor}{\cancel{#1}}% math-mode: use \cancel
		\else
			\textcolor{deletedcolor}{\sout{#1}}% text-mode: use \sout
		\fi
	\fi
}

\newboolean{pinapp}
\setboolean{pinapp}{true} % Set to true for proofs in appendix, false for proofs in main text

\newcommand{\changetheoremcounter}[1]{\setcounter{theorem}{\numexpr\getrefnumber{#1}-1\relax}}

\newcommand{\definetheorem}[4]{
	\expandafter\newcommand\csname theorem#2\endcsname{
		\begin{#1}
			\ifthenelse{\boolean{inappendix}}{}{\label{th_#2}}
			#3
		\end{#1}
	}
	\expandafter\newcommand\csname proof#2\endcsname{
		\begin{proof}
			#4
		\end{proof}
	}
}

\newcommand{\writetheorem}[1]{
	\csname theorem#1\endcsname
	\ifthenelse{\boolean{pinapp}}{}
	{\csname proof#1\endcsname}
}

\newcommand{\calltheorem}[1]{
	\changetheoremcounter{th_#1}
	\csname theorem#1\endcsname
	\csname proof#1\endcsname
}

\newcommand{\myequation}[2]{%
	\begin{equation}
		#2
		\ifthenelse{\boolean{inappendix}}{\nonumber}{\label{eq_#1}}
	\end{equation}
}

\newboolean{inappendix}
\setboolean{inappendix}{false}

\usepackage{lastpage}
\begin{document}

\title{Bias-variance decompositions: the exclusive privilege of Bregman divergences}
\author{\name Tom Heskes \email tom.heskes@ru.nl \\
\addr Institute for Computing and Information Sciences\\
Radboud University\\
Nijmegen, The Netherlands}

%\editor{Mladen Kolar}

\maketitle

\begin{abstract}%
Bias-variance decompositions are widely used to understand the generalization performance of machine learning models. While the squared error loss permits a straightforward decomposition, other loss functions -- such as zero-one loss or $L_1$ loss -- either fail to sum bias and variance to the expected loss or rely on definitions that lack the essential properties of meaningful bias and variance. Recent research has shown that clean decompositions can be achieved for the broader class of Bregman divergences, with the cross-entropy loss as a special case. However, the necessary and sufficient conditions for these decompositions remain an open question.

In this paper, we address this question by studying continuous, nonnegative loss functions that satisfy the identity of indiscernibles \added{(zero loss if and only if the two arguments are identical)}\deleted{under mild regularity conditions}. We prove that so-called $g$-Bregman or rho-tau divergences are the only such loss functions that have a clean bias-variance decomposition. A $g$-Bregman divergence can be transformed into a standard Bregman divergence through an invertible change of variables. This makes the squared \deleted{Mahalanobis distance}\added{Euclidean distance}, up to such a variable transformation, the only symmetric loss function with a clean bias-variance decomposition. Consequently, common metrics such as zero-one and $L_1$ losses cannot admit a clean bias-variance decomposition, explaining why previous attempts have failed. We also examine the impact of relaxing the restrictions on the loss functions and how this affects our results.
\end{abstract}

%\begin{keywords}
%	Bias-variance decomposition, Bregman divergence
%\end{keywords}

\section{Introduction}

Bias-variance decompositions\deleted{,} \added{have been} introduced \added{to the machine learning community} by \citet{geman1992biasvariance}\deleted{,}\added{. They} decompose the expected loss over a family of machine learning models into a bias term, depending on \deleted{the}\added{a suitable} central prediction \added{(defined more specifically below, and corresponding to the expected prediction in the case of squared loss)}, and a variance term, independent of the (distribution of the) actual labels. These decompositions are often used to study and better understand the generalization performance of machine learning approaches, also in more recent work that questions the standard adagium that bias decreases and variance increases as a function of model complexity \citep{neal2018modern,belkin2019reconciling}.

The bias-variance decomposition for the squared error loss has a very appealing form and is also relatively easy to derive. However, generalizing this decomposition to other loss functions, such as the zero-one loss or the $L_1$ loss, has proven to be extremely difficult, if not impossible \citep{wolpert1997bias,domingos2000unified,james2003variance}. Attempts to define bias and variance for these other loss functions often result in decompositions that either do not sum to the expected loss, or require relaxing requirements, such as allowing the bias to depend on the distribution of predictions or the variance to depend on the label \citep{domingos2000unified,james2003variance}.

Fortunately, there are loss functions beyond the squared error that do have clean bias-variance decompositions. For instance, bias-variance decompositions have been derived for loss functions related to the log likelihood of probabilistic models, such as the cross-entropy loss \citep{heskes1998bias}. More recent work has shown that these clean decompositions apply to the entire class of Bregman divergences, which includes the squared error and cross-entropy loss as special cases \citep{pfau2013generalized,gupta2022ensembles,wood2023unified}. However, it is still unclear whether these are the only loss functions with such decompositions. As noted in recent literature, ``the necessary and sufficient conditions for such a [clean bias-variance] decomposition are an open research question" \citep{brown2024biasvariance}.

In this paper, we aim to answer this question. \deleted{At least partially, since we do put some mild restrictions on the loss functions under consideration: they must be}\added{We focus on loss functions that are} nonnegative\deleted{, sufficiently differentiable almost everywhere,} and satisfy the identity of indiscernibles \added{(zero loss if and only if the two arguments are identical)}. These restrictions are specified in Section~\ref{sec_loss}. In Section~\ref{sec_biasvar}, we discuss the bias-variance decomposition and what we, following earlier work \citep{neal2018modern}, refer to as a `clean' bias-variance decomposition. Bregman divergences, and their relation to probability distributions within the exponential family, are introduced in Section~\ref{sec_bregman}. Here we also \deleted{define}\added{consider} a slight generalization of the standard Bregman divergence, the so-called $g$-Bregman divergence \added{\citep{nishiyama2018generalized} or rho-tau divergence  \citep{zhang2004divergence,naudts2018rho}}, which can be turned into a standard Bregman divergence through an invertible change of variables. Section~\ref{sec_unique} proves our main contribution, which is reflected in less subtle form in the title of this paper: \deleted{under mild regularity conditions, }$g$-Bregman divergences are the only nonnegative loss functions with the identity of indiscernibles that have a clean bias-variance decomposition. In Section~\ref{sec_relax} we discuss the extent to which we can relax the restrictions on the loss functions and how this affects our results. \ifthenelse{\boolean{pinapp}}{All main proofs can be found in the appendix.}{} The examples presented throughout this paper are intended to illustrate the ideas in the main text.

\section{Loss Functions}\label{sec_loss}

We consider loss functions $\loss: \labelspace \times \labelspace \rightarrow \mathbb{R}_{\geq 0}$\added{, not necessarily symmetric,} where $\loss(\vlab,\vpred)$ measures the error when predicting $\vpred$ for the actual label (target) $\vlab$. We assume that $\vlab$ and $\vpred$ belong to the same domain $\labelspace \subseteq \mathbb{R}^d$, which is a subset of the $d$-dimensional Euclidean space and treat $\vlab$ and $\vpred$ as $d$-dimensional column vectors.

\added{Note that our notation slightly deviates from recent machine learning literature, where $y$ often denotes the labels and $\hat{y} = f(x)$ the predictions. There the dependency of the prediction on the model $f$ and inputs $x$ is made explicit, but for the analysis here this dependency is irrelevant: what matters is the distribution over predictions (and possibly labels), not its origin. Using $y$ for predictions allows concise notation such as $\bar{y}$ for central predictions and avoids overloading $f$.}

Unless specified otherwise, we will assume that $\loss$ is nonnegative ($\loss(\vlab,\vpred) \geq 0$), satisfies the identity of indiscernibles ($\loss(\vlab,\vpred) = 0$ if and only if $\vlab = \vpred$), and is continuous everywhere.

\begin{boxedexample}{ex_squared}
	Arguably the best known example of such a loss function is the squared Euclidean distance
	\begin{equation}
		\loss(\vlab,\vpred) = \|\vlab-\vpred\|^2 = \sum_{i=1}^d |\lab_i - 	\pred_i|^2 \: .
	\label{eq_euclidean}
	\end{equation}
	Its domain $\labelspace$ is typically the whole of $\mathbb{R}^d$. For $d=1$ it simplifies to the squared error
	\[
		\loss(\lab,\pred) = (t-y)^2 \: .
	\]
	
	If the labels $\lab_i$ and predictions $\pred_i$ can be interpreted as probabilities, the squared Euclidean distance becomes the Brier score. With $d$ probabilities, we can set $\labelspace$ to the probability simplex:
	\[
		\labelspace = \Delta^{d-1} = \left\{ \vpred \in [0,1]^d ~ \left| ~ \sum_{i=1}^d \pred_i = 1 \right. \right\} \: .
	\]
	When optimizing loss functions, we need to take into account such equality constraints. \added{An alternative to the Brier score is the cross-entropy loss or Kullback-Leibler divergence, described below in Example~\ref{ex_kldiv}.}\deleted{ An alternative approach is to redefine the loss function in terms of less variables, for example by expressing the last probability $y_d$ as a function of the others.}
\end{boxedexample}

\deleted{We further assume that the loss function $\loss$ is twice continuously differentiable with respect to $\vlab$ and continuously differentiable with respect to $\vpred$, almost everywhere, which we will indicate as $\loss \in C^{2,1}(\labelspace \times \labelspace\setminus S)$, where $S$ has Lebesgue measure 0. The `almost everywhere' is included to also take into consideration Minkowski-type loss functions as in the following example.}\added{The squared Euclidean distance is smooth and differentiable in both its arguments. It can be generalized to a broader class of Minkowski-type losses, as in the following example, which are not necessarily differentiable everywhere.}

\begin{boxedexample}{}
	We can generalize the squared Euclidean distance to loss functions of the form
	\[
		\loss(\vlab,\vpred) = \|\vlab-\vpred\|_\epsilon^{\epsilon} = \sum_{i=1}^d |\lab_i - \pred_i|^\epsilon \: .
	\]
	For $\epsilon = 2$, we get back the squared Euclidean loss and for $\epsilon = 1$ the $L_1$ norm.
	For $0 < \epsilon < 2$, this loss function is infinitely differentiable everywhere, except for the set $S = \{ (\vlab,
	\vpred) \in \labelspace \times \labelspace ~|~ \exists i \text{~such that~} \lab_i = \pred_i \}$.\deleted{ Since this set has measure zero, our criterion of being sufficiently differentiable `almost everywhere' is being satisfied.}
\end{boxedexample}

We explicitly do not require the loss function $\loss$ to be a metric: it need not be symmetric, nor does it need to satisfy the triangle inequality.

\section{Bias-Variance Decomposition}\label{sec_biasvar}

Suppose that we have some distribution of predictions $\vrpred$, for example resulting from machine learning models trained on different training sets and/or started from different initializations. We write $\expectation_{\vrpred}$ for the expectation with respect to such a distribution. This expectation can be taken over any relevant quantity, whether it involves a probability density over continuous predictions, a probability mass distribution over discrete predictions, or a finite ensemble of models generating predictions. For a fixed vector label $\vlab$, the expected squared Euclidean distance can be decomposed as
\[
	\underbrace{\expectation_{\vrpred} \|\vlab-\vrpred\|^2}_{\text{expected loss}} = \underbrace{\left\|\vlab - \expectation_{\vrpred} \vrpred\right\|^2}_{\text{bias}} + \underbrace{\expectation_{\vrpred} \left\| \expectation_{\vrpred} \vrpred - \vrpred\right\|^2}_{\text{variance}} \: .
\]
We will refer to the first term on the right-hand side as the bias and the second term as the variance. The important observation here is that the variance is independent of the label $\vlab$, whereas the bias only measures the error between the label and the \deleted{average}\added{expected} prediction. If we now take into account that labels $\vrlab$ are randomly drawn from some distribution \errata{independent of the predictions $\vrpred$,} and then also consider the expectation $\expectation_{\vrlab}$ with respect to this distribution, we can further decompose
\begin{equation}
	\underbrace{\expectation_{\vrlab,\vrpred} \|\vrlab - \vrpred\|^2}_{\text{expected loss}} = \underbrace{\expectation_{\vrlab} \left\|\vrlab - \expectation_{\vrlab} \vrlab\right\|^2}_{\text{intrinsic noise}} + \underbrace{\left\|\expectation_{\vrlab} \vrlab - \expectation_{\vrpred} \vrpred \right\|^2}_{\text{bias}} + \underbrace{\expectation_{\vrpred} \left\| \expectation_{\vrpred} \vrpred - \vrpred\right\|^2}_{\text{variance}} \: .
\label{eq_bvsquared}
\end{equation}
The first term on the right-hand side is referred to as the intrinsic noise or irreducible error: it provides a lower bound on the loss that can be obtained even with the best possible predictions.

Following~\citet{neal2018modern}, we refer to~(\ref{eq_bvsquared}) as a `clean' bias-variance decomposition that, for general loss functions, we define as follows.

\begin{definition}
	A loss function $\loss: \labelspace \times \labelspace \rightarrow \Rplus$ is said to have a \emph{clean} bias-variance decomposition if, for all valid distributions of labels $\vrlab$ and predictions $\vrpred$, the expected loss decomposes as
	\begin{equation}
		\underbrace{\expectation_{\vrlab,\vrpred} \loss(\vrlab,\vrpred)}_{\text{\rm expected loss}} = \underbrace{\expectation_{\vrlab} \loss(\vrlab,\vlabopt)}_{\text{\rm intrinsic noise}} + \underbrace{\loss(\vlabopt,\vpredopt)}_{\text{\rm bias}} + \underbrace{\expectation_{\vrpred} \loss(\vpredopt,\vrpred)}_{\text{\rm variance}} \: ,
	\label{eq_clean}
	\end{equation}
	with
	\begin{equation}
		\vlabopt = \argmin_{\vpred \in \labelspace} \expectation_{\vrlab} \loss(\vrlab,\vpred) \mbox{~~and~~} \vpredopt = \argmin_{\vlab \in \labelspace} \expectation_{\vrpred} \loss(\vlab,\vrpred) \: .
	\label{eq_optimal}
	\end{equation}
\end{definition}
In the decomposition~(\ref{eq_clean}), all three terms have a natural interpretation\added{. }\deleted{ in terms of}\added{The} irreducible error \deleted{(}\added{is }independent of the distribution of the predictions\deleted{)}, \added{the} bias \deleted{(}only depends on the distributions of the labels and the predictions through summary statistics\deleted{)}, and \added{the} variance \deleted{(}\added{is} independent of the distribution of the labels\deleted{), respectively}. Most importantly, these terms collectively sum to the expected loss. \citet{wood2023unified} refer to such bias-variance decompositions as `additive' decompositions. In this paper, we adopt the term `clean' to highlight that the definitions of bias and variance \deleted{should }have a natural interpretation, viewing additivity as an inherent requirement of any decomposition.

\deleted{The summary statistic}\added{By definition,} $\vlabopt$ is \deleted{defined to be }the prediction closest to the distribution of the labels and, vice versa, $\vpredopt$ the label closest to the distribution of the predictions. \added{Here ``closest'' is interpreted relative to the argument being optimized: $\vlabopt$ minimizes the loss over the first argument, while $\vpredopt$ minimizes the loss over the second argument.} In the literature on Bregman divergences, $\vlabopt$ and $\vpredopt$ are referred to as right(-sided) and left(-sided) centroid~\citep{brown2024biasvariance,nielsen2009sided}, or the central label and prediction~\citep{gupta2022ensembles}, respectively.

The specific ordering of $\vlabopt$ and $\vpredopt$ in~(\ref{eq_clean}) is intentional and important: any other ordering is doomed to fail for all asymmetric loss functions (and does not make a difference for symmetric loss functions).

\definetheorem{proposition}{ordering}{
	\deleted{The specific ordering of $\vrlab$, $\vlabopt$, $\vpredopt$, and $\vrpred$ on the right-hand side of~(\ref{eq_clean}) is the only ordering that can lead to a clean bias-variance decomposition for asymmetric loss functions $\loss$ with the identity of indiscernibles.}\added{For asymmetric loss functions $\loss$ with the identity of indiscernibles, the specific ordering of arguments in the loss terms on the right-hand side of~(\ref{eq_clean}) is the only one that yields a clean bias-variance decomposition.}
	}{
	Let us first consider interchanging $\vlabopt$ and $\vpredopt$ in $\loss(\vlabopt,\vpredopt)$ to try
	\[
		\expectation_{\vrlab,\vrpred} \loss(\vrlab,\vrpred) = 	\expectation_{\vrlab} \loss(\vrlab,\vlabopt) + \loss(\vpredopt,\vlabopt) + \expectation_{\vrpred} \loss(\vpredopt,\vrpred) \: .
	\]
	This \deleted{should}\added{must} also hold in the special case of an arbitrary single label $\vlab$ and arbitrary single prediction $\vpred$, which simplifies to
	\[
		\expectation_{\vrlab,\vrpred} \loss(\vrlab,\vrpred) = \loss(\vlab,\vpred) = \loss(\vlab,\vlabopt) + \loss(\vpredopt,\vlabopt) + \loss(\vpred,\vpredopt) = \loss(\vpred,\vlab) \: ,
	\]
	since, by identity of indiscernibles, $\vpredopt = \vpred$ and $\vlabopt = \vlab$. That is, the bias-variance decomposition with the ordering $\loss(\vpredopt,\vlabopt)$ instead of $\loss(\vlabopt,\vpredopt)$ forces $\loss$ to be symmetric, in contradiction with being appropriate for asymmetric loss functions.

	Similar reasoning rules out interchanging $\vrlab$ and $\vlabopt$ in the intrinsic noise term $\expectation_{\vrlab} \loss(\vrlab,\vlabopt)$. In case of a single prediction $\vpred = \vpredopt$, the bias-variance decomposition with a reverse intrinsic noise term would give
	\[
		\expectation_{\vrlab,\vrpred} \loss(\vrlab,\vrpred) = \expectation_{\vrlab} \loss(\vrlab,\vpredopt) = \expectation_{\vrlab} \loss(\vlabopt,\vrlab) + \loss(\vlabopt,\vpredopt) \: .
	\]
	This \deleted{should}\added{must} also apply when $\vpredopt$ happens to coincide with $\vlabopt$, which yields $\expectation_{\vrlab} \loss(\vrlab,\vlabopt) = \expectation_{\vrlab} \loss(\vlabopt,\vrlab)$ and \deleted{should}\added{must} be valid for any distribution of labels $\vrlab$. Again, we conclude that $\loss$ then must be symmetric for the bias-variance decomposition to uphold. The exact same reasoning applies to interchanging $\vpredopt$ and $\vrpred$ in $\expectation_{\vrpred} \loss(\vpredopt,\vrpred)$ \added{as well as to changing the order in two or even all three terms}.
	}

\writetheorem{ordering}

\deleted{Furthermore, for the decomposition~(\ref{eq_clean}) to work, $\vpredopt$ must be the unique minimizer of the variance term and $\vlabopt$ the unique minimizer of the intrinsic noise term. In other words, we may as well interpret their definition~(\ref{eq_optimal}) as a consequence of having a clean bias-variance decomposition rather than as a natural requirement for such a decomposition.}\added{Furthermore, for the decomposition~(\ref{eq_clean}) to hold, $\vpredopt$ must be the unique minimizer of the variance term and $\vlabopt$ the unique minimizer of the intrinsic noise term. Thus, their definition in~(\ref{eq_optimal}) is not merely a convenient requirement but in fact a necessary consequence of having a clean bias-variance decomposition.}

\definetheorem{proposition}{optimal}{
	\deleted{Suppose that the bias-variance decomposition~(\ref{eq_clean}) of a loss function $\loss$ with the identity of indiscernibles applies for some $\vlabopt$ independent of the distribution of the predictions $\vrpred$ and some $\vpredopt$ independent of the distribution of the labels $\vrlab$. Then $\vlabopt$ and $\vpredopt$ \deleted{should}\added{must} obey~(\ref{eq_optimal}), that is, they must be the central label and prediction, minimizing the intrinsic noise and the variance term, respectively. Furthermore, these minimizers are necessarily unique.}\added{Suppose the bias-variance decomposition~(\ref{eq_clean}) holds for a loss function $\loss$ with the identity of indiscernibles, using some $\vlabopt$ independent of the distribution of predictions $\vrpred$ and some $\vpredopt$ independent of the distribution of labels $\vrlab$. Then $\vlabopt$ and $\vpredopt$ must coincide with the central label and central prediction from~(\ref{eq_optimal}), minimizing the intrinsic noise and variance terms, respectively. Moreover, these minimizers are necessarily unique.}
	}{
	To prove that $\vpredopt$ must be the central prediction, we consider the special case of a single vector label $\vlab$ and suppose that the decomposition holds for some $\tilde{\vpred}$, that is,
	\[
		\expectation_{\vrpred} \loss(\vlab,\vrpred) = \loss(\vlab,\tilde{\vpred}) + \expectation_{\vrpred} \loss(\tilde{\vpred},\vrpred) \: .
	\]
	Substituting $\vlab = \vpredopt$, with $\vpredopt$ by definition the label minimizing the variance term, we get
	\[
		0 \leq \loss(\vpredopt,\tilde{\vpred}) = \expectation_{\vrpred} 	\loss(\vpredopt,\vrpred) - \expectation_{\vrpred} \loss(\tilde{\vpred},\vrpred) \leq 0 \: ,
	\]
	where the first inequality follows from the nonnegativity of the loss function, and the second inequality follows from $\vpredopt$ minimizing the variance term. By the identity of indiscernibles, $\loss(\vpredopt,\tilde{\vpred}) = 0$ implies $\tilde{\vpred} = \vpredopt$. This argument and the outcome $\vpredopt = \tilde{\vpred}$ apply, even when $\tilde{\vpred}$ is a minimizer of the variance term. Hence the variance term must have a unique minimizer. The proof for the central label $\vlabopt$ follows the exact same steps, simply interchanging the roles of predictions and labels.
	}

\writetheorem{optimal}

The uniqueness of the minimizer of the variance term results from the loss function's clean bias-variance decomposition, rather than assumptions like (quasi-)convexity. Similarly, we do not \deleted{feel the }need to specify whether the domain $\labelspace$ is convex, open, or closed: the existence of a clean bias-variance decomposition and hence of the bias $\loss(\vlabopt,\vpredopt)$ implies that the central label and central prediction must be within the same domain $\labelspace$. If they are not, we can try to extend the domain $\labelspace$ such that they are.

\section{Bregman Divergences}\label{sec_bregman}

Bregman divergences~\citep{bregman1967relaxation} form a specific class of loss functions and have become quite popular in the recent machine learning literature \citep{siahkamari2020learning,cilingir2020deep,amid2019robust}. The Bregman divergence $\bregman{A}: \labelspace \times \labelspace \rightarrow \Rplus$, with $\labelspace$ a convex subset of $\mathbb{R}^d$, is fully determined by its generating function $A: \labelspace \rightarrow \mathbb{R}$:
\begin{equation}
	\bregman{A}(\vlab,\vpred) = (\vpred - \vlab)^\top \nabla_{\vpred} A(\vpred) - A(\vpred) + A(\vlab)  \: .
\label{eq_bregman}
\end{equation}
$A$ is strictly convex and differentiable. A Bregman divergence is strictly convex in its first argument, but not necessarily in its second argument~\citep{bauschke2001convexity}. \citet{banerjee2005clustering} provide many other properties as well as examples of Bregman divergences.

\begin{boxedexample}{}
	The squared Euclidean distance~(\ref{eq_euclidean}) from Example~\ref{ex_squared} follows by taking $A(\vpred) = \|\vpred\|^2$. Generalizing this to $A(\vpred) = \vpred^\top K \vpred$ for a positive definite matrix $K$, we obtain the squared Mahalanobis distance
	\[
		\loss(\vlab,\vpred) = (\vlab-\vpred)^\top K (\vlab-\vpred) \: .
	\]
	The squared Mahalanobis distance is the only symmetric Bregman divergence: all other Bregman divergences are asymmetric~\citep{boissonnat2010bregman}. 
\end{boxedexample}

Bregman divergences relate to distributions in the exponential family as follows. In canonical form, any distribution in the exponential family can be written
\[
	p_{\vpred,A}(\vect{z}) = h(\vect{z}) \exp[\vpred^\top \vphi(\vect{z}) - A(\vpred)] \: ,
\]
with $h(\vect{z})$ the base function, $\vphi(\vect{z})$ the sufficient statistics, $A(\vpred)$ the log-partition function, and $\vpred$ the canonical \added{(or natural)} parameters. A Bregman divergence $\bregman{A}$ between parameters $\vlab$ and $\vpred$ equals the Kullback-Leibler divergence between two distributions in the exponential family with log-partition function $A$ and canonical parameters $\vpred$ and $\vlab$ \citep[for example,][]{chowdhury2023bregman}:
\[
	\mbox{KL}(p_{\vpred,A},p_{\vlab,A}) = (\vpred - \vlab)^\top \nabla_{\vpred} A(\vpred) - A(\vpred) + A(\vlab) = \bregman{A}(\vlab,\vpred) \: ,
\]
where we used
\[
	\expectation_{\vect{Z}} \vphi(\vect{Z}) = \nabla_{\vpred} A(\vpred) \: .
\]

\begin{boxedexample}{ex_gauss}
	A univariate Gaussian distribution with mean $m_\pred$ and variance $\variance_\pred$ can be written in the form
	\[
		p_{\vpred,A}(z) = \frac{1}{\sqrt{2 \pi \variance_\pred}} \exp \left[ - \frac{1}{2 \variance_\pred} (m_\pred-z)^2 \right] = h(z) \exp \left[ \vpred^\top \vphi(z) - A(\vpred) \right] \: ,		
	\]
	by choosing $\vpred = (m_\pred/\variance_\pred,-1/(2\variance_\pred))$, $h(z) = 1$, $\vphi(z) = (z,z^2)$, and
	\[
		A(\vpred) = - \frac{\pred_1^2}{4 \pred_2}
		\deleted{+}\added{-} \frac{1}{2} \log \left( - \frac{y_2}{\pi}\right) \: ,
	\]
	where $\pred_1 \in \mathbb{R}$ and $\pred_2 \in \mathbb{R}_{<0}$.
	Plugging this into the definition of the Bregman divergence, we obtain for the KL divergence between two Gaussian distributions with canonical parameters $\vpred$ and $\vlab$ the intruiging form
	\begin{equation}
		\loss(\vlab,\vpred) = \added{-} \frac{1}{2} + \frac{\pred_1 \lab_1}{2 \pred_2} - \frac{\pred_1^2 \lab_2}{4 \pred_2^2} \deleted{-}\added{+} \frac{\lab_2}{2 \pred_2} - \frac{\lab_1^2}{4 \lab_2} \deleted{+}\added{-} \frac{1}{2} \log \left( \frac{\lab_2}{\pred_2} \right) \: ,
	\label{eq_bgausst}
	\end{equation}
	which may look more familiar when translated back to means and variances:
	\begin{equation}
		\loss(\vlab,\vpred) = \frac{(m_\pred - m_\lab)^2}{2\variance_\lab} \deleted{+}\added{-} \frac{1}{2} \left[             1 - \frac{\variance_\pred}{\variance_\lab} + \log \left(\frac{\variance_\pred}{\variance_\lab} \right) \right] \: .
	\label{eq_bgaussm}
	\end{equation}
\end{boxedexample}

Another, more direct connection between the exponential family and Bregman divergences goes through the log likelihood. Consider a member of the exponential family with canonical parameter $\tilde{\vpred}$, sufficient statistic $\vphi(\vect{z})$, and log-partition function $B(\tilde{\vpred})$. Suppose that our model outputs an estimate of the canonical parameter $\tilde{\vpred}$, and receives observation $\vect{z}$. Then a natural, and often used loss function to measure the quality of our estimate is minus the log likelihood, which, up to irrelevant constants independent of $\tilde{\vpred}$, reads
\begin{equation}
	- \log p_{B}(\vect{z};\tilde{\vpred}) = - \tilde{\vpred}^\top \vphi(\vect{z}) + B(\tilde{\vpred}) \added{+ \; \text{const}} \: .
\label{eq_loglikelihood}
\end{equation}
This loss function does not yet satisfy our requirements of nonnegativity and identity of indiscernibles. However, we can turn it into one by rewriting the observation $\vect{z}$ in terms of the sufficient statistic $\vlab = \vphi(\vect{z})$ and (Legendre) transforming the canonical parameter $\tilde{\vpred}$:
\[
	\vpred = \nabla_{\tilde{\vpred}} B(\tilde{\vpred}) \: ,
\]
and, correspondingly, $B$ to its convex conjugate
\[
	A(\vpred) = \tilde{\vpred}^\top \nabla_{\tilde{\vpred}} B(\tilde{\vpred}) - B(\tilde{\vpred}) \: .
\]
such that
\[
	\tilde{\vpred} = \nabla_{\vpred} B(\vpred) \mbox{~~and~~} B(\tilde{\vpred}) = \vpred^\top \nabla_{\vpred} A(\vpred) - A(\vpred) \: .
\]
Substituting these into~(\ref{eq_loglikelihood}), we obtain, again up to constants independent of $\tilde{\vpred}$ and $\vpred$,
\begin{equation}
	- \log p_{B}(\vect{z};\tilde{\vpred}) = (\vpred-\vlab)^\top \nabla_{\vpred} A(\vpred) - A(\vpred) \added{+ \; \text{const}} \: .
\label{eq_almost}
\end{equation}
By adding $A(\vlab)$, we arrive at the Bregman divergence $\bregman{A}(\vlab,\vpred)$ from~(\ref{eq_bregman}).

\begin{boxedexample}{ex_kldiv}
	Let us consider the case of probabilistic binary classification with Bernouilli likelihood function
	\[
		p(\lab;\pred) = \pred^\lab (1-\pred)^{1-\lab} \: ,
	\]
	with $\lab \in \{0,1\}$ and $\pred \in [0,1]$. Taking the negative log likelihood, we obtain the cross-entropy loss
	\[
		- \log p(\lab;\pred) = -\lab \log \pred - (1 - \lab) \log (1-\pred) \: \deleted{.}\added{,}
	\]
	\added{with the convention $0 \log 0 = 0$. Thus, when $\pred \in \{0,1\}$, the loss is finite if $\lab = \pred$, and infinite otherwise.}
	This is indeed of the form~(\ref{eq_almost}) if we set $A(\pred) = \pred \log \pred + (1-\pred) \log (1-\pred)$. The cross-entropy loss by itself may not be a Bregman divergence, but can be turned into one by adding $A(\lab)$ to arrive at the Kullback-Leibler divergence
	\[
		\loss(\lab,\pred) = \lab \log \left(\frac{\lab}{\pred}\right) + (1 - \lab) \log \left(\frac{1-\lab}{1 - \pred} \right) \: ,
	\]
	which works fine for any $\lab \in [0,1]$ and is\deleted{, with convention $0 \log 0 = 0$,} equal to the cross-entropy loss at the extremes $\lab = 0$ and $\lab = 1$.
	
	Generalizing this to the multinomial, multi-class case, for $\vlab,\vpred \in \Delta^{d-1}$ (see Example~\ref{ex_squared}), we obtain the KL divergence
	\begin{equation}
		\loss(\vlab,\vpred) = \sum_{i=1}^d \lab_i \log \left(\frac{\lab_i}{\pred_i}\right) + \sum_{i=1}^d \pred_i - \sum_{i=1}^d t_i \: ,
	\label{eq_kldiv}
	\end{equation}
	which is a Bregman divergence with generating function $A(\vpred) = \sum_{i=1}^d \pred_i \log \pred_i - \sum_{i=1}^d \pred_i$. Adding  $\sum_{i=1}^d y_i$ - $\sum_{i=1}^d t_i$ to the loss function makes the Kullback-Leibler divergence a proper Bregman divergence that satisfies~(\ref{eq_bregman})\deleted{ and}\added{. Expressed in this form, the KL divergence} guarantees the identity of indiscernibles also when we extend $\labelspace$ from the probability simplex $\Delta^{d-1}$ to the \deleted{whole of $\Rplus^d$}\added{unit cube $[0,1]^d$}, that is, when we ignore the equality constraint $\sum_i \pred_i = 1$:
	\[
		\argmin_{\vpred \in \Delta^{d-1}} \loss(\vlab,\vpred) = \argmin_{\vpred \in [0,1]^d} %\deleted{\Rplus^{d-1}}\added{[0,1]^d}}
		\loss(\vlab,\vpred) = \vlab \mbox{~~and~~} \argmin_{\vlab \in \Delta^{d-1}} \loss(\vlab,\vpred) = \argmin_{\vlab \in
		%\deleted{\Rplus^{d-1}}\added{[0,1]^d}
		[0,1]^d} \loss(\vlab,\vpred) = \vpred \: .
	\]
	We will refer to\deleted{~(\ref{eq_kldiv})} \added{a Bregman divergence that has the property that the identity of indiscernibles still holds when we ignore equality constraints} as a Bregman divergence in proper form.
\end{boxedexample}

For now, we will assume that the observations $\vect{z}$ and the sufficient statistics/labels $\vlab$ (and hence parameters/predictions $\vpred$) have the same dimensionality. In Section~\ref{sec_relax}, we examine scenarios where the observation $\vect{z}$ has a lower dimensionality than the label $\vlab$.

We \deleted{define}\added{consider} a slightly generalized variant of the Bregman divergence, by \citet{nishiyama2018generalized} referred to as the $g$-Bregman divergence \added{and by~\citet{naudts2018rho} and \citet{brekelmans2024variational} as the rho-tau divergence}. We write $\vectg(\labelspace)$ for the image of $\labelspace$ under mapping $\vectg$ and define the set $\labelspace \setminus \labelspace_{\text{eq}}$ as the original $\labelspace$, but then without taking into account any (linear) equality constraints. \added{So, in Example~\ref{ex_kldiv} above, $\labelspace$ refers to the probability simplex $\Delta^{d-1}$, $\labelspace_{\text{eq}}$ is the hyperplane $\{\pred \in \labelspace : \sum_i \pred_i = 1\}$ enforcing the equality constraint, and $\labelspace \setminus \labelspace_{\text{eq}}$ then corresponds to the $d$-dimensional cube $[0,1]^d$.}

\begin{definition}
	Given an invertible (bijective) function $\vectg: \labelspace \rightarrow \vectg(\labelspace)$ and strictly convex generating function $A: \vectg(\labelspace) \rightarrow \mathbb{R}$, with $\labelspace$ and $\vectg(\labelspace \setminus \labelspace_{\text{\rm eq}})$ both convex subsets of $\mathbb{R}^d$, the $g$-Bregman divergence $\gbregman{A}{\vectg}: \labelspace \times \labelspace$ is defined as
	\begin{equation}
		\gbregman{A}{\vectg}(\vlab,\vpred) = \bregman{A}(\vectg(\vlab),\vectg(\vpred)) = (\vectg(\vpred) - \vectg(\vlab))^\top \nabla_{\vectg(\vpred)} A(\vectg(\vpred)) - A(\vectg(\vpred)) + A(\vectg(\vlab)) \: ,
	\label{eq_generalized}
	\end{equation}
\end{definition}

Loosely speaking, a $g$-Bregman divergence $\gbregman{A}{\vectg}$ between $\vlab$ and $\vpred$ is a standard Bregman divergence between $\tilde{\vlab} = \vectg(\vlab)$ and $\tilde{\vpred} = \vectg(\vpred)$. It can be interpreted as the Kullback-Leibler divergence between two distributions in the exponential family in noncanonical form, with canonical parameter function $\vectg$ and log-partition function $A \circ \vect{g}$. A $g$-Bregman divergence need not be convex in its first, nor in its second argument. It is nonnegative and by construction satisfies the identity of indiscernibles.

\begin{boxedexample}{ex_reverse}
	For $\vlab, \vpred \in \Delta^{d-1}$, the reverse KL divergence,
	\begin{equation}
		\loss(\vlab,\vpred) = \sum_{i=1}^d \pred_i \log \left( \frac{\pred_i}{\lab_i} \right) + \sum_{i=1}^d \lab_i - \sum_{i=1}^d \pred_i \: , 
	\label{eq_reverse}
	\end{equation}
	is a $g$-Bregman divergence with mapping $\vectg$ such that $g_i(\vpred) = \log \pred_i$ and $A(\vectg) = \sum_{i=1}^d \exp (g_i)$.
	
	\deleted{We can further generalize this to}\added{The forward KL divergence~(\ref{eq_kldiv}) and the reverse KL divergence~(\ref{eq_reverse}) can be considered special cases of} the so-called $\alpha$-divergence\added{~\citep{amari1985differential}}
	\begin{equation}
		\loss(\vlab,\vpred) = - \frac{1}{\alpha (1-\alpha)}  \sum_{i=1}^d \lab_i^\alpha \pred_i^{1-\alpha} + \frac{1}{1-\alpha} \sum_{i=1}^d \lab_i + \frac{1}{\alpha} \sum_{i=1}^d \pred_i ,
	\label{eq_alpha}
	\end{equation}
	\added{where we follow the convention of \citet{zhu1995information}.}
	\deleted{which, f}\added{F}or $0 < \alpha < 1$, \added{this $\alpha$-divergence} is a $g$-Bregman divergence with mapping $\vectg$ such that $g_i(\vpred) = \frac{1}{1-\alpha} \pred_i^\alpha$ and $A(\vectg) = (1-\alpha)^{(1-\alpha)/\alpha}\sum_{i=1}^d g_i^{1/\alpha}$. \added{The forward KL divergence~(\ref{eq_kldiv}) is obtained in the limit $\alpha \rightarrow 1$ and the reverse KL divergence~(\ref{eq_reverse}) in the limit $\alpha \rightarrow 0$.}
\end{boxedexample}

\definetheorem{proposition}{reverse}{
	The reverse of a $g$-Bregman divergence, with $\vlab$ and $\vpred$ interchanged, is another $g$-Bregman divergence:
	\[
		\gbregman{A}{\vectg}(\vlab,\vpred) = \gbregman{B}{\vectf}(\vpred,\vlab) \mbox{~~with~~} \vectf(\vpred) = \nabla_{\vectg(\vpred)} A(\vectg(\vpred)) \mbox{~~and~~} B(\vectf(\vpred)) = \vectg(\vpred)^\top \vectf(\vpred) - A(\vectg(\vpred)) \: .
	\]
	}{
	\added{See Theorem~6 of~\cite{zhang2004divergence}. }A $g$-Bregman divergence is a sum of three terms, each depending on both $\vpred$ and $\vlab$, only on $\vpred$, or only on $\vlab$:
	\begin{eqnarray*}
		\gbregman{A}{\vectg}(\vlab,\vpred) & = & - \underbrace{\vectg(\vlab)^\top \nabla_{\vectg(\vpred)} A(\vectg(\vpred))}_{\text{both $\vpred$ and $\vlab$}} + \underbrace{\vectg(\vpred)^\top \nabla_{\vectg(\vpred)} A(\vectg(\vpred)) - A(\vectg(\vpred))}_{\text{only $\vpred$}}
		+ \underbrace{A(\vectg(\vlab))}_{\text{only $\vlab$}} \\
		& = & - \underbrace{\vectf(\vpred)^\top \nabla_{\vectf(\vlab)} 	B(\vectf(\vlab))}_{\text{both $\vpred$ and $\vlab$}} + \underbrace{B(\vectf(\vpred))}_{\text{only $\vpred$}} + \underbrace{\vectf(\vlab)^\top \nabla_{\vectf(\vlab)} B(\vectf(\vlab)) - B(\vectf(\vlab))}_{\text{only $\vlab$}} \\
		& = & \gbregman{B}{\vectf}(\vpred,\vlab)\: .
	\end{eqnarray*}
	Equating the first term depending on both $\vpred$ and $\vlab$ leads to $\vectf(\vpred) = \nabla_{\vectg(\vpred)} A(\vectg(\vpred))$. Equating the second term only depending on $\vpred$ gives $B(\vectf(\vpred)) = \vectg(\vpred)^\top \vectf(\vpred) - A(\vectg(\vpred))$. The other terms precisely match with the reverse mapping from $\{B,\vectf\}$ to $\{A,\vectg\}$.
	}

\writetheorem{reverse}

With $\gbregman{A}{\vectg}(\vlab,\vpred) = \gbregman{B}{\vectf}(\vpred,\vlab)$, we can write the $g$-Bregman divergence in the concise form
\begin{equation}
	\gbregman{A}{\vectg}(\vlab,\vpred) = A(\vectg(\vlab)) - \vectf(\vpred)^\top \vectg(\vlab) + B(\vectf(\vpred)) \: .
\label{eq_concise}
\end{equation}
\added{This form is sometimes referred to as the mixed parameterization or canonical form~\citep{brekelmans2024variational} and is reminiscent of the Fenchel-Young divergence~\citep{blondel2020learning}.}
$A$ and $B$ are each other's convex conjugates, in the sense that for all $\vpred \in \labelspace$,
\begin{equation}
	A^*(\vectf(\vpred)) = \sup_{\vlab \in \labelspace} \vectg(\vlab)^\top \vectf(\vpred) - A(\vectg(\vlab)) = B(\vectf(\vpred)) \: .
\label{eq_conjugate}
\end{equation}
We will therefore refer to $\{A,\vectg\}$ and $\{B,\vectf\}$ as each other's dual pair. The duality~(\ref{eq_conjugate}) still applies for $\vpred \in \labelspace$ when equality constraints are involved, but then we are no longer guaranteed that $A^*(\vectf) = B(\vectf)$ for arbitrary $\vectf$. The form~(\ref{eq_concise}) is very convenient to directly read off the dual pairs $\{A,\vectg\}$ and $\{B,\vectf\}$ from any $g$-Bregman divergence in proper form.

Given a $g$-Bregman divergence $\gbregman{A}{\vectg}$ with dual pair $\{B,\vectf\}$, we will call $\vectg(\vlab)$ the label $\vlab$ in canonical (natural) form and $\vectf(\vpred)$ the prediction $\vpred$ in moment (expectation) form. We define
\[
	\bvlab = \vectg^{-1} \left( \expectation_{\vrlab} \vectg(\vrlab) \right) \mbox{~~and~~} \bvpred = \vectf^{-1} \left( \expectation_{\vrpred} \vectf(\vrpred) \right) \: ,
\]
the $\vectg$-mean label and $\vectf$-mean prediction, respectively, that follow by averaging the labels in canonical form and predictions in moment form and transforming the resulting averages back to the original space. These means play a role in the next lemmas and theorem, which demonstrate that, in the absence of equality constraints, all $g$-Bregman divergences have a clean bias-variance decomposition.

\definetheorem{lemma}{means}{
	Consider a $g$-Bregman divergence $\gbregman{A}{\vect{g}}: \labelspace \times \labelspace \rightarrow \Rplus$ with dual pair $\{B,\vect{f}\}$. If the $\vectg$-mean label $\bvlab$ satisfies any constraints $\labelspace$ may impose, that is, if $\bvlab \in \labelspace$, then it minimizes the intrinsic noise term:
	\[
		\bvlab = \vlabopt = \argmin_{\vpred \in \labelspace} \expectation_{\vrlab} \gbregman{A}{\vectg}(\vrlab,\vpred) \: ,
	\]
	yielding
	\[
		\text{\rm intrinsic noise} = \expectation_{\vrlab} 	\gbregman{A}{\vectg}(\vrlab,\vlabopt) = \expectation_{\vrlab} A(\vectg(\vrlab)) - A(\vectg(\vlabopt)) \: .
	\]
	Along the same lines, if the $\vectf$-mean prediction $\bvpred$ satisfies any constraints $\labelspace$ may impose, that is, if $\bvpred \in \labelspace$, then it minimizes the variance term:
	\myequation{momaveraging}{
		\bvpred = \vpredopt = \argmin_{\vlab \in \labelspace} 	\expectation_{\vrpred} \gbregman{A}{\vectg}(\vlab,\vrpred) \: ,
	}
	yielding 
	\[
		\text{\rm variance} = \expectation_{\vrpred} 	\gbregman{B}{\vectf}(\vrpred,\vpredopt) = \expectation_{\vrpred} B(\vectf(\vrpred)) - B(\vectf(\vpredopt)) \: .
	\]
	}{
	We follow the same line of reasoning as \added{in Proposition~1 of} \citet{banerjee2005clustering} for a standard Bregman divergence \added{and as in Theorem~3 of \citet{brekelmans2024variational} for the rho-tau divergence}. Let $\{B,\vectf\}$ be the dual pair of $\{A,\vectg\}$. For any $\tvlab \in \labelspace$, we consider the difference
	\begin{eqnarray*}
		\expectation_{\vrlab} \gbregman{A}{\vectg}(\vrlab,\tvlab) - \expectation_{\vrlab} \gbregman{A}{\vectg}(\vrlab,\bvlab) & = & 
		B(\vectf(\tvlab)) - B(\vectf(\bvlab)) + (\vectf(\bvlab) - \vectf(\tvlab))^\top \expectation_{\vrlab} \vectg(\vrlab) \\
		& = & B(\vectf(\tvlab)) - B(\vectf(\bvlab)) + (\vectf(\bvlab) - \vectf(\tvlab))^\top \vectg(\bvlab) \\
		& = & \gbregman{B}{\vectf}(\bvlab,\tvlab) \geq 0\: .
	\end{eqnarray*}
	That is, when the $\vect{g}$-mean $\bvlab$ indeed exists and satisfies any constraints $\labelspace$ may impose, there can be no other $\tvlab \in \labelspace$ that leads to a lower intrinsic noise. Plugging $\vlabopt = \bvlab$ into the definition of the intrinsic noise, we get
	\begin{eqnarray*}
		\text{intrinsic noise} & = & \expectation_{\vrlab} \gbregman{A}{\vectg}(\vrlab,\bvlab) \\
		& = & \left(\vectg(\bvlab) - \expectation_{\vrlab} \vectg(\vrlab)\right)^\top  \nabla_{\vectg(\bvlab)} A(\vectg(\bvlab)) + \expectation_{\vrlab} A(\vectg(\vrlab)) - A(\vectg(\bvlab)) \\
 		& = & \expectation_{\vrlab} A(\vectg(\vrlab)) - A(\vectg(\bvlab)) \: .
 	\end{eqnarray*}
	
	The proof for the $\vectf$-mean prediction $\bvpred$ follows the exact same lines, interchanging labels for predictions and $\{A,\vectg\}$ for $\{B,\vectf\}$.
	}

\writetheorem{means}

\definetheorem{lemma}{bvgbregman}{
	Consider a $g$-Bregman divergence $\gbregman{A}{g}: \labelspace \times \labelspace \rightarrow \Rplus$ with generating function $A: \vectg(\labelspace) \rightarrow \mathbb{R}$ and invertible mapping $\vectg: \labelspace \rightarrow \vectg(\labelspace)$. Assume that the central label $\vlabopt$ minimizing the intrinsic noise term is equal to the $\vectg$-mean label $\vectg^{-1}(\expectation_\vrlab \vectg(\vrlab))$, and that the central prediction $\vpredopt$ minimizing the variance term is equal to the $\vectf$-mean prediction $\vectf^{-1}(\expectation_\vrpred \vectf(\vrpred))$, where $\vectf(\vpred) = \nabla_{\vectg(\vpred)} A(\vectg(\vpred))$. Then $\gbregman{A}{\vectg}$ has a clean bias-variance decomposition:
	\myequation{bvgbregman1}{
		\underbrace{\expectation_{\vrlab,\vrpred} \gbregman{A}{\vectg}(\vrlab,\vrpred)}_{\text{\rm expected loss}} = \underbrace{\expectation_{\vrlab} \gbregman{A}{\vectg}(\vrlab,\vlabopt)}_{\text{\rm intrinsic noise}} + \underbrace{\gbregman{A}{\vectg}(\vlabopt,\vpredopt)}_{\text{\rm bias}} + \underbrace{\expectation_{\vrpred} \gbregman{A}{\vectg}(\vpredopt,\vrpred)}_{\text{\rm variance}} \: .
	}
    }{
	The proof follows by making the change of variables from $\vlab$ and $\vpred$ to $\vectg(\vlab)$ and $\vectg(\vpred)$ in the bias-variance decomposition for (standard) Bregman divergences provided by \citet{pfau2013generalized} and \citet{gupta2022ensembles}.

	For completeness and illustration, we here provide an alternative proof that makes use of the symmetry following from Proposition~\ref{th_reverse}. With $\{B,\vectf\}$ the dual pair of $\{A,\vectg\}$, taking expectations in the symmetric form~(\ref{eq_concise}) yields
	\begin{eqnarray*}
		\lefteqn{\expectation_{\vrlab,\vrpred} 	\gbregman{A}{\vectg}(\vrlab,\vrpred) = \expectation_{\vrlab} A(\vectg(\vrlab)) - \expectation_{\vrpred} \vectf(\vrpred)^\top \expectation_{\vrlab} \vectg(\vrlab) + \expectation_{\vrpred} B(\vectf(\vrpred))} \\
		& \! \! \! \! \! = \underbrace{\expectation_{\vrlab} A(\vectg(\vrlab)) - A(\vectg(\vlabopt))}_{\text{intrinsic noise}} + \underbrace{A(\vectg(\vlabopt)) - \vectf(\vpredopt)^\top \vectg(\vlabopt) + B(\vectf(\vpredopt))}_{\text{bias}} + \underbrace{\expectation_{\vrpred} B(\vectf(\vrpred)) - B(\vectf(\vpredopt))}_{\text{variance}} \: .
	\end{eqnarray*}
	}

\writetheorem{bvgbregman}

\begin{boxedexample}{ex_gaussagain}
	\added{To illustrate the bias-variance decomposition of a $g$-Bregman divergence, we return to}\deleted{In} Example~\ref{ex_gauss}\deleted{, }. Here we derived a \added{(standard) }Bregman divergence between the parameters of two univariate Gaussian distributions. The loss function~(\ref{eq_bgausst}) has a quite peculiar form, essentially because for a Bregman divergence we need to fix a parameterization such that the additive term in the loss function that depends on both $\vlab$ and $\vpred$ is linear in $\vlab$.
	
	The more natural form~(\ref{eq_bgaussm}) can be viewed as a $g$-Bregman divergence between parameters $\vlab = (m_\lab,\variance_\lab)$ and $\vpred = (m_\pred,\variance_\pred)$ with
	\[
		\vectg(m,\variance) = \left(\frac{m}{\variance},-\frac{1}{2 	\variance}\right) \mbox{~~and~~} \vectf(m,\variance) = (m,m^2 + \variance) \: .
	\]
	\added{In this parameterization, the bias-variance decomposition is relatively straightforward to derive.}
	The central label follows from averaging the canonical parameters and transforming these back to moment form:
	\[
		\opt{\variance}_\lab = \left( \expectation_{\rvariance_\lab} 	\left(\rvariance_\lab^{-1}\right) \right)^{-1} \mbox{~~and~~} \opt{m}_\lab = \opt{\variance}_\lab \expectation_{M_\lab,\rvariance_\lab} \left(\frac{M_\lab}{\rvariance_\lab}\right) \: ,
	\]
    To arrive at the central prediction we need to average the moment parameters
	\[
		\opt{m}_\pred = \expectation_{M_\pred} M_\pred \mbox{~~and~~} 	\opt{\variance}_\pred = \expectation_{M_\pred,\rvariance_\pred} \left( M_\pred^2 + \rvariance_\pred \right) - {\opt{m}_\pred}^2 \: .
	\]
	The bias-variance decomposition then reads
	\[
		\underbrace{\expectation_{\vrlab,\vrpred} 	\loss(\vrlab,\vrpred)}_{\text{expected loss}}
		= \underbrace{\expectation_{\vrlab} \tilde{A}(\vrlab) - 	\tilde{A}(\vlabopt)}_{\text{intrinsic noise}} + \underbrace{\loss(\vlabopt,\vpredopt)}_{\text{bias}} + \underbrace{\expectation_{\vrpred} \tilde{B}(\vrpred) - \tilde{B}(\vpredopt)}_{\text{variance}} \: ,
	\]
	with
	\[
		\tilde{A}(\vlab) = A(\vectg(m_\lab,\variance_\lab)) = 	\frac{m_\lab^2}{2 \variance_\lab} \deleted{-}\added{+} \frac{1}{2} \log \added{(2 \pi} \variance_\lab \added{)} \mbox{~and~} \tilde{B}(\vpred) = B(\vectf(m_\pred,\variance_\pred)) = \added{-} \frac{1}{2} \log \added{(2 \pi} \variance_\pred\added{)} \deleted{+}\added{-} \frac{1}{2} \: .
	\]
\end{boxedexample}

\definetheorem{theorem}{gbregmanineq}{
	Consider a $g$-Bregman divergence $\gbregman{A}{g}: \labelspace \times \labelspace \rightarrow \Rplus$ with generating function $A: \vectg(\labelspace) \rightarrow \mathbb{R}$ and invertible mapping $\vectg: \labelspace \rightarrow \vectg(\labelspace)$, and assume that $\vectg(\labelspace) = \text{\rm dom}(A)$ is a convex feasible set defined by inequality constraints only\errata{, and that its dual image $\vectf(\labelspace) = \nabla A(\vectg(\labelspace))$ is also convex}. Then this $g$-Bregman divergence has a clean bias-variance decomposition of the form~(\ref{eq_bvgbregman1}).
	}{
	Since $\vectg$ is invertible and $\vectg(\labelspace)$ is assumed to be convex with only inequality constraints (if any), we can pull the expectation and the optimization through the mapping and do not need to worry about any equality constraints:
	\begin{eqnarray*}
		\vlabopt = \argmin_{\vpred \in \labelspace} \expectation_{\vrlab} 	\gbregman{A}{\vectg}(\vrlab,\vpred) = \vectg^{-1} \left(\argmin_{\vectg(\vpred) \in \vectg(\labelspace)} \expectation_{\vectg(\vrlab)} \bregman{A}(\vectg(\vrlab),\vectg(\vpred)) \right) =  \vectg^{-1} \left( \expectation_{\vrlab} \vectg(\vrlab) \right) = \bvlab \: .
	\end{eqnarray*}
	With $\{B,\vectf\}$ the dual pair of $\{A,\vectg\}$, $B$ is the convex conjugate of $A$\deleted{ and its domain $\vectf(\labelspace)$ is a convex set. The}\errata{. Given the assumption that $\vectf(\labelspace)$ is a convex set,} the exact same reasoning then applies to the central prediction $\vpredopt$, by interchanging labels for predictions and $\{A,\vectg\}$ for $\{B,\vectf\}$:
	\[
		\vpredopt = \argmin_{\vlab \in \labelspace} \expectation_{\vrpred} \gbregman{A}{\vectg}(\vlab,\vrpred) = \vectf^{-1} \left( \expectation_{\vrpred} \vectf(\vrpred) \right) = \bvpred \: .
	\]
	Since the conditions of Lemma~\ref{th_bvgbregman} are now satisfied, this $g$-Bregman divergence has a clean bias-variance decomposition.
	}
	
\writetheorem{gbregmanineq}

When $\labelspace = \text{dom}(\vectg)$ includes equality constraints, the situation becomes considerably more complicated. If $\vectg$ is a nonlinear function, the linear constraints in $\labelspace$ turn into nonlinear constraints in $\vectg(\labelspace)$. Consequently, this transformation can lead to a nonconvex optimization problem, as the feasible region defined by the nonlinear constraints may no longer be convex. We therefore first consider the case of a standard Bregman divergence, that is, with $\vectg$ the identity function. 

\definetheorem{theorem}{bregconstr}{
	Consider a Bregman divergence $\bregman{A}: \labelspace \times \labelspace \rightarrow \Rplus$ with generating function $A: \labelspace \rightarrow \mathbb{R}$, and dual pair $\{B,\vectf\}$. Assume that the feasible set $\labelspace$ contains equality constraints that can be written in the form $W \vpred = \vectb$, with $W \in \mathbb{R}^{\kappa \times d}$, $\vectb \in \mathbb{R}^\kappa$, and $\kappa$ the number of constraints\errata{, and that the dual image of the unconstrained domain, $\vectf(\labelspace \setminus \labelspace_{\text{\rm eq}}) = \nabla A(\vectg(\labelspace \setminus \labelspace_{\text{\rm eq}}))$, is a convex set}. Then this Bregman divergence has a clean bias-variance decomposition of the form
	\[
		\underbrace{\expectation_{\vrlab,\vrpred} \bregman{A}(\vrlab,\vrpred)}_{\text{\rm expected loss}} = \underbrace{\expectation_{\vrlab} A(\vrlab) - A(\vlabopt)}_{\text{\rm intrinsic noise}} + \underbrace{\bregman{A}(\vlabopt,\vpredopt)}_{\text{\rm bias}} + \underbrace{\vlambda^\top \vectb + \expectation_{\vrpred} B(\vectf(\vrpred)) - B(\vectf(\vpredopt))}_{\text{\rm variance}} \: ,
	\]
	where
	\[
		\vlabopt = \expectation_{\vrlab} \vrlab \mbox{~~and~~}
		\vpredopt = \vectf^{-1} \left( \expectation_{\vrpred} \vectf(\vrpred) + \vlambda^\top W \right) \mbox{~~with $\vlambda$ such that~~} W \vpredopt = \vectb \: .
	\]
	}{
	The central label is defined as
	\[
		\vlabopt = \argmin_{\vpred \in \labelspace} \expectation_{\vrlab} \bregman{A}(\vrlab,\vpred) = \argmin_{\vpred \in \labelspace} \left[B(\vectf(\vpred)) - \vectf(\vpred)^\top \expectation_{\vrlab} \vrlab \right] \: .
	\]
	Adding Lagrange multipliers $\vlambda \in \mathbb{R}^\kappa$ to incorporate the linear constraints $W \vpred - \vectb = 0$, and setting the derivative with respect to $\vectf(\vpred)$ to zero, while making use of $\nabla_{\vectf(\vpred)} B(\vectf(\vpred)) = \vpred$, we obtain
	\[
		\vlabopt = \expectation_{\vrpred} \vrpred + \vlambda^\top W \: .
	\]
	With all labels $\vrpred \in \labelspace$, so does $\vlabopt$ for $\vlambda = \vect{0}$, yielding the usual expression
	\[
		\text{intrinsic noise} = \expectation_{\vrlab} \bregman{A}(\vrlab,\vlabopt) = \expectation_{\vrlab} A(\vrlab) - A(\vlabopt) \: .
	\]
	
	When solving for the central prediction, the constraints do matter. We have
	\[
		\vpredopt = \argmin_{\vlab  \in \labelspace} \expectation_{\vrpred} \bregman{A}(\vlab,\vrpred) = \argmin_{\vlab \in \labelspace} \left[A(\vlab) - \vlab^\top \expectation_{\vrpred} \vectf(\vrpred)\right] \: .
	\]
	Adding Lagrange multipliers $\vlambda$ and setting the derivative with respect to $\vlab$ to zero, while making use of $\nabla_\vlab A(\vlab) = \vectf(\vlab)$, we obtain
	\[
		\vectf(\vpredopt) = \expectation_{\vrpred} \vectf(\vrpred) + 	\vlambda^\top W \: .
	\]
	\errata{Since $\vectf(\labelspace \setminus \labelspace_{\text{\rm eq}})$ is convex by assumption, the right-hand side is guaranteed to lie in the range of $\vectf$, ensuring that a unique feasible $\vpredopt = \vectf^{-1}\left(\expectation_{\vrpred} \vectf(\vrpred) + \vlambda^\top W\right)$ exists.}
	Here $\vlambda$ \deleted{should}\added{must} be chosen such that indeed $W \vpredopt = \vectb$. Plugging this solution in the expression for the variance, we arrive at
	\begin{eqnarray*}
		\text{variance} & = & \expectation_{\vrpred} 	\gbregman{B}{\vectf}(\vrpred,\vpredopt) = \vlambda^\top W \vpredopt + \expectation_{\vrpred} B(\vectf(\vrpred)) - B(\vectf(\vpredopt)) \\
		& = & \vlambda^\top \vectb + \expectation_{\vrpred} B(\vectf(\vrpred)) - B(\vectf(\vpredopt)) \: .
	\end{eqnarray*}
	}

\writetheorem{bregconstr}

\deleted{Given the complete symmetry between $\vlab$ and $\vpred$ in our analysis, everything that applies to a Bregman divergence, should also apply to a reverse Bregman divergence.}
\added{Because of the symmetry between $\vlab$ and $\vpred$ in our analysis, every statement about Bregman divergences applies equally to reverse Bregman divergences. This leads to the following theorem.}

\begin{corollary}
	Consider a reverse Bregman divergence, that is, a $g$-Bregman divergence $\gbregman{A}{g}: \labelspace \times \labelspace \rightarrow \Rplus$ with generating function $A: \labelspace \rightarrow \mathbb{R}$, and dual pair $\{B,\vectf\}$ such that $\vectf(\vpred) = \vpred$ \errata{and $\vectg(\vlab) = \nabla_{\vlab} B(\vlab)$}. Assume that the feasible set $\labelspace$ contains equality constraints that can be written in the form $W \vpred = \vectb$, with $W \in \mathbb{R}^{\kappa \times d}$, $\vectb \in \mathbb{R}^\kappa$, and $\kappa$ the number of constraints\errata{, and that the image of the unconstrained domain, $\vectg(\labelspace \setminus \labelspace_{\text{\rm eq}})$, is a convex set}. Then this $g$-Bregman divergence has a clean bias variance decomposition of the form
	\[
		\underbrace{\expectation_{\vrlab,\vrpred} 	\gbregman{A}{g}(\vrlab,\vrpred)}_{\text{\rm expected loss}} = \underbrace{\expectation_{\vrlab} A(\vectg(\vrlab)) - A(\vectg(\vlabopt)) + \vlambda^\top \vectb}_{\text{\rm intrinsic noise}} + \underbrace{\gbregman{A}{g}(\vlabopt,\vpredopt)}_{\text{\rm bias}} + \underbrace{ \expectation_{\vrpred} B(\vrpred) - B(\vpredopt)}_{\text{\rm variance}} \: ,
	\]
	where
	\[	
		\vlabopt = \vectg^{-1} \left( \expectation_{\vrlab} \vectg(\vrlab) + \vlambda^\top W \right) \mbox{~~with $\vlambda$ such that~~} W \vlabopt = \vectb \mbox{~,~~and~~} \vpredopt = \expectation_{\vrpred} \vrpred \: .
	\]
\end{corollary}

The proof follows directly by interchanging the roles of $\vlab$ and $\vpred$ in Theorem~\ref{th_bregconstr}.

\begin{boxedexample}{}
	Let us revisit the KL divergence~(\ref{eq_kldiv}) from Example~\ref{ex_kldiv}. This is a (standard) Bregman divergence, so $\vectg(\vlab) = \vlab$, with generating function $A(\vectg) = \sum_{i=1}^d g_i \log g_i - \sum_{i=1}^d g_i$. Its dual pair $\{B,\vectf\}$ is such that $f_i(\vpred) = \log \pred_i$ and $B(\vectf) = \sum_{i=1}^d \exp(f_i)$. With $\vones$ the $d$-dimensional column vector of all ones, $W = \vones^\top$, and $b = 1$, the conditions for Theorem~\ref{th_bregconstr} are satisfied and we get the bias-variance decomposition, in correspondence with~\citet{heskes1998bias},
	\[
		\underbrace{\expectation_{\vrlab,\vrpred} \sum_{i=1}^d \rlab_i \log 	\left( \frac{\rlab_i}{\rpred_i} \right)}_{\text{expected loss}} = \underbrace{\expectation_{\vrlab} \sum_{i=1}^d \rlab_i \log \rlab_i - \sum_{i=1}^d \opt{\rlab}_i \log \opt{\rlab}_i}_{\text{\rm intrinsic noise}} + \underbrace{\sum_{i=1}^d \opt{\rlab}_i \log \left( \frac{\opt{\rlab}_i}{\opt{\rpred}_i} \right)}_{\text{\rm bias}} \underbrace{\vphantom{\sum_{i=1}^d} \errata{-} \log Z}_{\text{variance}} \: ,
	\]
	with
	\[
		\opt{\lab}_i = \expectation_{\vrlab} \rlab_i ~,~~ \opt{\pred}_i = 	\frac{1}{Z} \exp \left( \expectation_{\vrpred} \log \rpred_i \right) \mbox{~,~and~~} Z = \sum_{i=1}^d \exp \left(\expectation_{\vrpred} \log \rpred_i \right) \: .
	\]
	We obtain the bias-variance decomposition for the reverse KL divergence~(\ref{eq_reverse}) from Example~\ref{ex_reverse} by interchanging the role of labels and predictions: 
	\[
		\underbrace{\expectation_{\vrlab,\vrpred} \sum_{i=1}^d \rpred_i \log \left( \frac{\rpred_i}{\rlab_i} \right)}_{\text{expected loss}} = \underbrace{\vphantom{\sum_{i=1}^d} \errata{-} \log Z}_{\text{intrinsic noise}}  + \underbrace{\sum_{i=1}^d \opt{\pred}_i \log \left( \frac{\opt{\pred}_i}{\opt{\lab}_i} \right)}_{\text{\rm bias}} + \underbrace{\expectation_{\vrpred} \sum_{i=1}^d \rpred_i \log \rpred_i - \sum_{i=1}^d \opt{\pred}_i \log \opt{\pred}_i}_{\text{\rm variance}} \: ,
	\]
	with
	\[
		\opt{\pred}_i = \expectation_{\vrpred} \rpred_i ~,~~ \opt{\lab}_i =	\frac{1}{Z} \exp \left( \expectation_{\vrlab} \log \rlab_i \right) \mbox{~,~and~~} Z = \sum_{i=1}^d \exp \left(\expectation_{\vrlab} \log \rlab_i \right) \: .
	\]
	So, where for a standard KL divergence the central prediction follows from taking the geometric mean, the central prediction for the reverse KL divergence is the arithmetic mean. \added{For both the forward and the reverse KL divergence, we still arrive at a clean bias-variance decomposition when incorporating the normalization constraint on the probabilities.}
	
	\added{The $\alpha$-divergence~(\ref{eq_alpha}) for $0 < \alpha < 1$ is not a standard Bregman divergence, but can be interpreted as a $g$-Bregman divergence. Without the normalization constraint on the probabilities, so for $\labelspace = [0,1]^d$, it admits a clean bias-variance decomposition. However, when incorporating the normalization constraint, so for $\labelspace = \Delta^{d-1}$, this is no longer the case.} The minimizers of the intrinsic noise term and the variance term\deleted{ for the $\alpha$-divergence~(\ref{eq_alpha}) from Example~\ref{ex_reverse}} follow from taking the power mean
	\[
		\opt{\lab}_i = \frac{\left(\expectation_{\vrlab} 	\rlab_i^\alpha\right)^{1/\alpha}}{\sum_{j=1}^d \left(\expectation_{\vrlab} \rlab_j^\alpha\right)^{1/\alpha}} \mbox{~~and~~} \opt{\pred}_i = \frac{\left(\expectation_{\vrpred} \rpred_i^{1-\alpha}\right)^{1/(1-\alpha)}}{\sum_{j=1}^d \left(\expectation_{\vrpred} \rpred_j^{1-\alpha}\right)^{1/(1-\alpha)}} \: ,
	\]
	\deleted{Despite this relatively simple form, it is hard to see how to arrive at a clean bias-variance decomposition.}\added{We may still plug these into our definitions for the intrinsic noise, bias, and variance on the right-hand side of~(\ref{eq_clean}), but these then do not add up to the expected loss on the left-hand side.} An alternative \deleted{approach}\added{attempt} could be to explicitly \deleted{include}\added{incorporate} the \added{normalization }constraint and for example write
	\[
		\loss(\vlab,\vpred) = - \frac{1}{\alpha(1-\alpha)} \left( 	\sum_{i=1}^{d-1} \lab_i^\alpha \pred_i^{1-\alpha} + \phi(\vlab)^\alpha \phi(\vpred)^{1-\alpha} \right) + \frac{1}{\alpha (1-\alpha)} \: ,
	\]
	\deleted{with $\Sigma(\vlab) = \sum_{i=1}^{d-1} \lab_i$.}\added{with $\phi(\vlab) = 1- \sum_{i=1}^{d-1} \lab_i$ and where now $\labelspace = \left\{\vpred \in [0,1]^{d-1} \left| \sum_i y_i \leq 1 \right.\right\}$.} However, this resulting loss function no longer has the form of a $g$-Bregman divergence \added{and then also does not admit a clean bias-variance decomposition}.	
\end{boxedexample}

In summary, $g$-Bregman divergences have a clean bias-variance decomposition in the absence of equality constraints \errata{under suitable convexity conditions}. This property extends to both (standard) Bregman and reverse Bregman divergences, even when linear equality constraints are present. Our results are a generalization of those provided by \citet{pfau2013generalized} and \citet{gupta2022ensembles} for standard Bregman divergences without equality constraints. Beyond relatively simple cases -- such as those involving \deleted{easily }reversible linear transformations $\vectg$ or $\vectf$ -- it appears challenging to identify other 
$g$-Bregman divergences that \deleted{maintain}\added{admit} a clean bias-variance decomposition under equality constraints.

\section{Uniqueness}\label{sec_unique}

In the previous section, we have introduced $g$-Bregman divergences with a clean bias-variance decomposition. In this section we will show that $g$-Bregman divergences in fact are the only (nonnegative, with the identity of indiscernibles) loss functions that can have this property.

To simplify the analysis, we will for now ignore the distribution over labels and consider the case of a single, fixed label vector $\vlab$, for which we \deleted{should then have}\added{get}
\begin{equation}
	\underbrace{\expectation_{\vrpred} \loss(\vlab,\vrpred)}_{\text{expected loss}} = \underbrace{\loss(\vlab,\vpredopt)}_{\text{bias}} + \underbrace{\expectation_{\vrpred} \loss(\vpredopt,\vrpred)}_{\text{variance}} \: ,
\label{eq_singlelabel}
\end{equation}
for some $\vpredopt$ dependent on the distribution of the labels, but independent of the label $\vlab$. In the following lemma, we show that the second derivative of loss functions that satisfy~(\ref{eq_singlelabel}) factorizes in a special manner. Our line of reasoning is inspired by the analysis of \citet{banerjee2005optimality}.

\newcommand{\interior}{\text{\rm int}}

\definetheorem{lemma}{factorization}{
	Consider a loss function $\loss: \labelspace \times \labelspace \rightarrow \Rplus$\deleted{, with $\loss \in C^{2,1}(\labelspace \times \labelspace \setminus S)$,} that has a clean bias-variance decomposition. Then \errata{on the interior $\interior(\labelspace) \times \interior(\labelspace)$,} its \added{mixed }second derivative w.r.t.\ \deleted{label $\vlab$ and prediction $\vpred$, $\hessian_{\lab,\pred}(\vlab,\vpred) = \nabla_{\vlab} \nabla_{\vpred} \loss(\vlab,\vpred)$, necessarily factorizes as
	\myequation{decomposition}{
		\hessian_{\lab,\pred}(\vlab,\vpred) = H_1(\vlab) H_2^\top(\vpred) \: ,
	}}\added{prediction $\vpred$ and label $\vlab$, $\hessian_{\pred,\lab}(\vlab,\vpred) = \nabla_{\pred} \nabla_{\lab} \loss(\vlab,\vpred)$, necessarily factorizes as
	\myequation{decomposition}{
			\hessian_{\pred,\lab}(\vlab,\vpred) = H_2(\vpred) H_1^\top(\vlab) \: ,
	}}
	for some matrix functions $H_1: \labelspace \rightarrow \mathbb{R}^{d \times d}$ and $H_2: \labelspace \rightarrow \mathbb{R}^{d \times d}$ for any $(\vlab,\vpred) \in \labelspace \times \labelspace\deleted{ \setminus S}$.
	}{
	At first, we do not worry about potential irregularities and assume that the loss function $\loss$ is \deleted{twice continuously} differentiable with respect to the label $\vlab$ and \deleted{continuously differentiable with respect to }the prediction $\vpred$ across the \deleted{whole}\errata{interior} domain \deleted{$\labelspace \times \labelspace$}\errata{$\interior(\labelspace) \times \interior(\labelspace)$}. \added{We will justify below that this regularity is in fact enforced by the requirement that the clean bias-variance decomposition holds everywhere.}
	
	Let us consider a fixed label $\vlab \errata{\in \interior(\labelspace)}$ and a distribution with just two labels, $\vpred_1, \vpred_2 \errata{\in \interior(\labelspace)}$, that have probability $p$ and $1-p$, respectively. Our line of reasoning is sketched in Figure~\ref{fig_sketch} for $p = 1/2$.
	
	The central prediction $\vpredopt$ reads
	\[
		\vpredopt = \argmin_{\vlab \in \labelspace} p \loss(\vlab,\vpred_1) + (1-p) \loss(\vlab,\vpred_2) \: ,
	\]
	which, since $\loss$ is \deleted{continuously }differentiable with respect to $\vlab$, implies
	\[
		p \gradient_{\lab}(\vpredopt,\vpred_1) + (1-p) 	\gradient_{\lab}(\vpredopt,\vpred_2) = \vect{0} \: ,
	\]
	with convention $[\gradient_{\lab}(\vlab,\vpred)]_i = \nabla_{t_i} \loss(\vlab,\vpred)$. Now we slightly change to $\vpred_1' = \vpred_1 + \veps_1$ and $\vpred_2' = \vpred_2 + \veps_2$ with $\veps_1$ and $\veps_2$ constructed such that $\vpredopt$ stays the same. That is, for infinitesimal $\veps_1$ and $\veps_2$, we enforce the constraint
	\begin{eqnarray}
		\vect{0} & = & p \gradient_{\lab}(\vpredopt,\vpred_1 + \veps_1) + (1-p) \gradient_{\lab}(\vpredopt,\vpred_2 + \veps_2) \nonumber \\
		& = & p \gradient_{\lab}(\vpredopt,\vpred_1) + p \hessian_{\pred,\lab}(\vpredopt,\vpred_1)^\top \veps_1 + (1-p) \gradient_{\lab}(\vpredopt,\vpred_2) + (1-p) \hessian_{\pred,\lab}(\vpredopt,\vpred_2)^\top \veps_2 \nonumber \\
		& = & p \hessian_{\pred,\lab}(\vpredopt,\vpred_1)^\top \veps_1 + (1-p) \hessian_{\pred,\lab}(\vpredopt,\vpred_2)^\top \veps_2 \: ,
		\label{eq_samecentral}
	\end{eqnarray}
	with $[\hessian_{\pred,\lab}(\vlab,\vpred)]_{j,i} = \nabla_{\pred_j} \nabla_{\lab_i} \loss(\vlab,\vpred)$.
	
	If we indeed have a clean bias-variance decomposition, the bias is the expected loss minus the variance, which \deleted{should}\added{must} be the same before and after the transformation:
	\begin{eqnarray*}
		p \loss(\vlab,\vpred_1 + \veps_1) + (1-p) \loss(\vlab,\vpred_2 + \veps_2) - p \loss(\vpredopt,\vpred_1 + \veps_1) - (1-p) \loss(\vpredopt,\vpred_2 + \veps_2) = ~~~\\ = p \loss(\vlab,\vpred_1) + (1-p) \loss(\vlab,\vpred_2) - p \loss(\vpredopt,\vpred_1) - (1-p) \loss(\vpredopt,\vpred_2) \: .
	\end{eqnarray*}
	For infinitesimal $\veps_1$ and $\veps_2$, this boils down to
	\begin{equation}
		p (\gradient_{\pred}(\vlab,\vpred_1) - \gradient_{\pred}(\vpredopt,\vpred_1))^\top \veps_1 + (1-p) (\gradient_{\pred}(\vlab,\vpred_2) - \gradient_{\pred}(\vpredopt,\vpred_2))^\top \veps_2 = 0 \: ,
		\label{eq_constantbias}
	\end{equation}
	with $[\gradient_{\pred}(\vlab,\vpred)]_j = \nabla_{y_j} \loss(\vlab,\vpred)$.
	
	\deleted{Since we assume the loss function to be \deleted{continuously }differentiable with respect to both $\vlab$ and $\vpred$, we can apply Schwarz's theorem and use that the mixed partial derivatives are equal: $[\hessian_{\lab,\pred}(\vlab,\vpred)]_{i,j} = \nabla_{\lab_i} \nabla_{\pred_j} \loss(\vlab,\vpred) = \nabla_{\pred_j} \nabla_{\lab_i} \loss(\vlab,\vpred) = [\hessian_{\pred,\lab}(\vlab,\vpred)]_{j,i}$.}
	
	Combining~(\ref{eq_samecentral}) and (\ref{eq_constantbias}), we obtain the two equations
	\deleted{\[
		\begin{array}{ccccc}
			0 & = & p \hessian_{\lab,\pred}(\vpredopt,\vpred_1) \veps_1 & + & (1-p) \hessian_{\lab,\pred}(\vpredopt,\vpred_2) \veps_2 \\
			0 & = & p [\gradient_{\pred}(\vlab,\vpred_1) - \gradient_{\pred}(\vpredopt,\vpred_1)]^\top \veps_1 & + & (1-p) [\gradient_{\pred}(\vlab,\vpred_2) - \gradient_{\pred}(\vpredopt,\vpred_2)]^\top \veps_2 \: .
		\end{array}
	\]}
	\added{\[
	\begin{array}{ccccc}
		\vect{0} & = & p \hessian_{\pred,\lab}(\vpredopt,\vpred_1)^\top \veps_1 & + & (1-p) \hessian_{\pred,\lab}(\vpredopt,\vpred_2)^\top \veps_2 \\
		0 & = & p [\gradient_{\pred}(\vlab,\vpred_1) - \gradient_{\pred}(\vpredopt,\vpred_1)]^\top \veps_1 & + & (1-p) [\gradient_{\pred}(\vlab,\vpred_2) - \gradient_{\pred}(\vpredopt,\vpred_2)]^\top \veps_2 \: .
	\end{array}
	\]}
	The first equation tells us how to choose $\veps_2$ if we make an infinitesimal change $\veps_1$ and want $\vpredopt$ to stay the same. The second equation \deleted{should}\added{must} be valid for a loss function with a clean bias-variance decomposition, for any combination of $\veps_1$ and $\veps_2$ that satisfies the first one. When $\vlab$ is close to $\vpredopt$, that is, up to first order in the difference $\vlab-\vpredopt$, this second equation directly follows from the first. However, for loss functions with a clean bias-variance decomposition, this should not just hold for small differences between the label and the central prediction, but for any difference between the two.
	
	For the second equation to be valid, while still satisfying the first one, we \deleted{should}\added{must} be able to write the second equation as a linear combination of the first one. That is, there \deleted{should be}\added{must exist} some vector function $\vect{q}: \labelspace \times \labelspace \rightarrow \mathbb{R}^d$, which may still depend on $\vlab$ and $\vpredopt$, but cannot otherwise depend on $\vpred_1$ and $\vpred_2$, such that
	\deleted{\[
		\vect{q}(\vlab,\vpredopt)^\top \hessian_{\lab,\pred}(\vpredopt,\vpred_1)	= [\gradient_{\pred}(\vlab,\vpred_1) - \gradient_{\pred}(\vpredopt,\vpred_1)]^\top \: ,
	\]}
	\added{\[
		\hessian_{\pred,\lab}(\vpredopt,\vpred_1) \vect{q}(\vlab,\vpredopt)  = \gradient_{\pred}(\vlab,\vpred_1) - \gradient_{\pred}(\vpredopt,\vpred_1) \: ,
	\]}
	and the same with $\vpred_1$ replaced with $\vpred_2$. Since this \deleted{should}\added{must} hold for arbitrary $\vpred_1$, $\vpred_2$, and $p$, we may as well rewrite
	\begin{equation}
	\deleted{[\gradient_{\pred}(\vlab',\vpred) - gradient_{\pred}(\vlab,\vpred)]_j = \sum_k [\vect{q}(\vlab',\vlab)]_k [\hessian_{\lab,\pred}(\vlab,\vpred)]_{kj} \: ,
	}
	\added{[\gradient_{\pred}(\vlab',\vpred) - \gradient_{\pred}(\vlab,\vpred)]_j = \sum_k [\hessian_{\pred,\lab}(\vlab,\vpred)]_{jk} [\vect{q}(\vlab',\vlab)]_k \: ,
	}
	\label{eq_difdif}
	\end{equation}
	which \deleted{should}\added{must} be valid for some function $\vect{q}$ and arbitrary $\vlab'$, $\vlab$, and $\vpred$.
	
	\deleted{In words, a difference in $\gradient_{\pred}$ on the left-hand side, \deleted{should in a specific way be related}\added{has to be related in a specific way} to a derivative \deleted{in}\added{of} $\gradient_{\pred}$ on the right-hand side. For $\vlab'$ close enough to $\vlab$, in zeroth and first order in the difference $\vlab'-\vlab$, there are functions $\vect{q}$ that make this work: with $\vect{q}(\vlab,\vlab) = \vect{0}$ and $\left. \nabla_{\vlab'} \vect{q}(\vlab',\vlab) \right|_{\vlab'=\vlab} = \mathbb{I}_d$, where $\mathbb{I}_d$ is the identity matrix in $d$ dimensions, the left-hand side of~(\ref{eq_difdif}) equals the right-hand side up to first order in $\vlab'-\vlab$ for any sufficiently differentiable loss function $\loss$.
	
	To arrive at a constraint on loss functions with clean bias-variance decompositions, we therefore take this one order further and consider the second order in the difference $\vlab'-\vlab$. Twice taking derivatives w.r.t.\ $\vlab'$ on both sides and evaluating at $\vlab'=\vlab$, we obtain
	\[
		\nabla_{t_m} [\hessian_{\lab,\pred}(\vlab,\vpred)]_{ij} = \sum_k 	Q_{mik}(\vlab) [\hessian_{\lab,\pred}(\vlab,\vpred)]_{kj}  \: ,
	\]
	where we have defined
	\[
		Q_{mik}(\vlab) = \left. \nabla_{\lab'_m} \nabla_{\lab'_i} [\vect{q}(\vlab',\vlab)]_k \right|_{\vlab'=\vlab} \: .
	\]
	In matrix notation, with $[Q_m(\vlab)]_{ik} = Q_{mik}(\vlab)$, we have
	\[
		\nabla_{t_m} \hessian_{\lab,\pred}(\vlab,\vpred) = Q_m(\vlab) \hessian_{\lab,\pred}(\vlab,\vpred)  \: ,
	\]
	with solution
	\[
		\hessian_{\lab,\pred}(\vlab,\vpred) = \exp[\tilde{Q}(\vlab)] 	\tilde{R}(\vpred) \: ,
	\]
	for some matrices $\tilde{R}(\vpred)$ and $\tilde{Q}(\vlab)$ such that $\nabla_{\lab_m} \tilde{Q}(\vlab) = Q_m(\vlab)$. Setting $H_1(\vlab) = \exp[\tilde{Q}(\vlab)]$ and $H_2(\vpred) = \tilde{R}(\vpred)^\top$ gives the factorization~(\ref{eq_intermediate}).}
	
	\added{The left-hand side gives the difference in gradients $\gradient_{\pred}(\vlab',\vpred)-\gradient_{\pred}(\vlab,\vpred)\in\mathbb{R}^d$.
	The right-hand side expresses this vector as a linear combination of the columns of $\hessian_{\pred,\lab}(\vlab,\vpred)$, with coefficients	$\vect{q}(\vlab',\vlab)$ that depend only on the labels. For fixed $\vpred$ and	$\vlab$, the right-hand side therefore ranges precisely over the column space of
	$\hessian_{\pred,\lab}(\vlab,\vpred)$. Since~(\ref{eq_difdif}) must hold for arbitrary pairs of labels $\vlab'$ and $\vlab$ at fixed $\vpred$, every gradient
	difference $\gradient_{\pred}(\vlab',\vpred)-\gradient_{\pred}(\vlab,\vpred)$ must lie in this column space. Because this requirement holds for all choices of
	the reference label $\vlab$, the column space of $\hessian_{\pred,\lab}(\vlab,\vpred)$ cannot depend on $\vlab$.
		
	It follows that, for each fixed $\vpred$, there exists a matrix $H_2(\vpred)$ whose columns form a basis of this common column space, such that each column of
	$\hessian_{\pred,\lab}(\vlab,\vpred)$ lies in the span of $H_2(\vpred)$. Collecting the corresponding coefficients into a matrix $H_1(\vlab)$ yields the
	factorization
	\[
		\hessian_{\pred,\lab}(\vlab,\vpred) = H_2(\vpred)\,H_1(\vlab)^\top \: ,
	\]
	where $H_2(\vpred)$ encodes a fixed set of directions that span all possible gradient differences at prediction $\vpred$, and $H_1(\vlab)$ determines how these directions are combined for a given label $\vlab$.
		
	Since $\hessian_{\pred,\lab}(\vlab,\vpred)\in\mathbb{R}^{d\times d}$, its rank is at most $d$, so the factorization involves at most $d$ independent directions.
	Without loss of generality, we may therefore take $H_1,H_2\in\mathbb{R}^{d\times d}$ in~(\ref{eq_decomposition}). Choosing a larger	intermediate dimension would be redundant, while choosing a smaller one would	require $\hessian_{\pred,\lab}(\vlab,\vpred)$ to be rank-deficient everywhere.
	}
	
	\deleted{Throughout this proof, we assumed the set $S$ to be empty so that we did not need to worry about the existence of the derivatives. Backtracking the proof, we made Taylor expansions around the points $(\vpredopt,\vpred_1)$, $(\vpredopt,\vpred_2)$, $(\vlab,\vpred_1)$, and $(\vlab,\vpred_2)$. Since $\vpred_1$, $\vpred_2$, and $\vlab$ can be chosen arbitrarily, we can always make sure that all these points are outside of the set $S$ with Lebesgue measure zero. To arrive at the functional form for the second order derivative $\hessian_{\lab,\pred}$, we made a Taylor expansion around $(\vlab,\vpred)$, which makes the solution valid for any $(\vlab,\vpred) \notin S$.}\added{Finally, the differentiability assumptions made at the beginning of the proof are not just technical but are enforced by the clean bias-variance decomposition itself. The decomposition is an exact identity that must hold for all choices of labels and predictions, and therefore also under arbitrarily small perturbations of $\vpred_1$ and $\vpred_2$ that leave the central prediction $\vpredopt$ unchanged. Without local differentiability with respect to $\vlab$ and $\vpred$, the change of the loss under infinitesimal perturbations cannot, in general, be expressed linearly as in~(\ref{eq_samecentral}) and~(\ref{eq_constantbias}). Hence, for a clean bias-variance decomposition to hold locally, the loss must have well-defined first-order derivatives with respect to $\vlab$ and $\vpred$, so that the bias remains invariant under all perturbations preserving $\vpredopt$.
	}
}

\writetheorem{factorization}

\begin{figure}
	\begin{center}
		\vspace*{-1cm}
		\begin{tikzpicture}
			\coordinate (A) at (0,0);
			\coordinate (B) at (2,4);
			\coordinate (C) at (6,0);
			\coordinate (M) at ($(A)!0.5!(C)$);
			\draw[-Latex] (A) -- (M);
			\draw[-Latex] (C) -- (M);
			\draw[-Latex] (A) -- (B);
			\draw[-Latex] (C) -- (B);
			\draw[-Latex, dotted, thick] (M) -- (B);
			
			\node[below] at (A) {$\vpred_1$};
			\node[above] at (B) {$\vlab$};
			\node[below] at (C) {$\vpred_2$};
			\node[below] at (M) {$\vpredopt$};
			
			\coordinate (A2) at (-0.5,-0.5);
			\coordinate (C2) at (6.5,0.5);
			\draw[-Latex, dashed] (A2) -- (M);
			\draw[-Latex, dashed] (C2) -- (M);
			\draw[-Latex, dashed] (A2) -- (B);
			\draw[-Latex, dashed] (C2) -- (B);			
			% Add labels for the new triangle
			\node[left] at (A2) {$\vpred_1'$};
			\node[right] at (C2) {$\vpred_2'$};
			
			\node[above] at ($(A)!0.5!(M)$) {$v_1$};
			\node[below] at ($(C)!0.5!(M)$) {$v_2$};
			\node[below] at ($(A2)!0.5!(M)$) {$v_1'$};
			\node[above] at ($(C2)!0.5!(M)$) {$v_2'$};
			\node[right] at ($(B)!0.5!(M)$) {$b$};
			\node[right] at ($(A)!0.5!(B)$) {$l_1$};
			\node[left] at ($(C)!0.5!(B)$) {$l_2$};
			\node[left] at ($(A2)!0.5!(B)$) {$l_1'$};
			\node[right] at ($(C2)!0.5!(B)$) {$l_2'$};
			
		\end{tikzpicture}\\[-1cm]
	\end{center}
	\caption{Sketch of the construction used in the proof of Lemma~\ref{th_factorization}. We start from a distribution with just two predictions $\vpred_1$ and $\vpred_2$, here each with probability $1/2$, leading to the central prediction $\vpredopt$. \deleted{With e}\added{E}ach line segment \added{symbolically} represent\deleted{ing}\added{s} a loss\deleted{,} \added{--} for example, $v_1 = \loss(\vpredopt,\vpred_1)$ \deleted{and so on,}\added{-- with the arrow pointing from the second to the first argument. Specializing} the bias-variance decomposition\added{~(\ref{eq_singlelabel})} \deleted{implies}\added{to this two prediction case gives} $(l_1 + l_2)/2 = b + (v_1 + v_2)/2$. \added{The parallellogram indicated by the solid lines is essentially the same construction as in Figure~10 of~\citet{nielsen2021parallel}.}\\[-3mm] \\	Next, we consider a slight, infinitesimal change in the predictions, yielding $\vpred_1'$ and $\vpred_2'$, that keeps the central prediction $\vpredopt$ and hence the bias $b$ invariant. This invariance implies $(l_1' + l_2') - (v_1' + v_2') = 2 b = (l_1 + l_2) - (v_1 + v_2)$. \deleted{The requirement that this should hold for any such change that keeps the central prediction intact, also if this $\vpredopt$ is not necessarily close to the label $\vlab$, leads to a strong constraint on the form of the loss function's mixed second derivative.}\added{Requiring this to hold for any such change that keeps the central prediction intact -- even when $\vpredopt$ is not close to the label $\vlab$ -- places a strong constraint on the mixed second derivative of the loss function.}}
	\label{fig_sketch}
\end{figure}

The proof is given in the appendix and uses the construction sketched in Figure~\ref{fig_sketch}. We start from a distribution over labels $\vrpred$ that corresponds to some central prediction $\vpredopt$. Next we will make a tiny (infinitesimal) change in the distribution, which we deliberately choose such that $\vpredopt$ and hence the bias $\loss(\vlab,\vpredopt)$ stay the same. The expected loss and the variance will change, but, for loss functions with a clean bias-variance distribution, the difference between these two equals the bias and \deleted{should}\added{must} therefore be invariant under such a change for any choice of $\vlab$ and $\vpredopt$.

\deleted{Assuming that the loss function $\loss$ is continuously differentiable with respect to both $\vlab$ and $\vpred$, we can swap the order of differentation. It can then be easily shown that t}\added{T}he difference between the expected loss and the variance stays the same for any sufficiently smooth $\loss$, as long as $\vlab$ and $\vpredopt$ are (infinitesimally) close. However, the requirement that this invariance \deleted{should hold also}\added{also holds} for $\vlab$ and $\vpredopt$ further apart, imposes a serious constraint on the loss function. \deleted{Essentially making a second order Taylor expansion with respect to the difference between $\vlab$ and $\vpredopt$, we arrive at a differential equation for $\hessian_{\lab,\pred}(\vlab,\vpred)$, the solution of which must be of the form~(\ref{eq_decomposition}).}\added{In particular, invariance under all perturbations preserving $\vpredopt$ implies that differences of prediction gradients must lie in a fixed linear subspace determined by the mixed derivative of the loss. This enforces a separable structure of the mixed derivative and yields the factorization in~(\ref{eq_decomposition}).} \deleted{Backtracking the proof to check for the existence of the derivatives that are being used to arrive at the result, we conclude that the factorization applies for any $(\vlab,\vpred) \notin S$.}

With the help of Lemma~\ref{th_factorization}, we can now prove our main theorem.

\definetheorem{theorem}{bregman}{
	Consider a continuous, nonnegative loss function $\loss: \labelspace \times \labelspace \rightarrow \Rplus$\deleted{, with $\loss \in C^{2,1}(\labelspace \times \labelspace \setminus S)$,} that satifies the identity of indiscernibles and has a clean bias-variance decomposition. Then $\loss$ must be a $g$-Bregman divergence $\gbregman{A}{\vectg}$ of the form~(\ref{eq_generalized}) for some invertible mapping $\vectg: \labelspace \rightarrow \vectg(\labelspace)$ and strictly convex generating function $A: \vectg(\labelspace) \rightarrow \mathbb{R}$.
	}{
	From the factorization~(\ref{eq_decomposition}) in Lemma~\ref{th_factorization}, we know that the second derivative of the loss function $\loss$ with respect to $\vlab$ and $\vpred$, $\hessian_{\lab,\pred}(\vlab,\vpred)$, \deleted{almost everywhere }factorizes as
	\[
	\deleted{[\hessian_{\lab,\pred}(\vlab,\vpred)]_{ij} = \sum_k [H_1(\vlab)]_{ik} 	[H_2(\vpred)]_{jk} = [H_1(\vlab) H_2(\vpred)^\top]_{ij} \: ,}
	\added{[\hessian_{\pred,\lab}(\vlab,\vpred)]_{ij} = \sum_k [H_2(\vpred)]_{ik} 	[H_1(\vlab)]_{jk} = [H_2(\vpred) H_1(\vlab)^\top]_{ij} \: ,}
	\]
	for some \added{$d \times d$} matrices $H_1(\vlab)$ and $H_2(\vpred)$. To arrive at the loss function itself, we perform two integrations, first with respect to $\vlab$ and then with respect to $\vpred$. \deleted{While doing so, we can ignore potential singularities where $\hessian_{\lab,\pred}$ is undefined or infinite, because these singularities form a set $S$ of Lebesgue measure zero. According to the Lebesgue differentiation theorem, the contribution of such sets to the integral is negligible, ensuring that the integration process remains valid and $\loss$ remains continuous.}
	
	Integration with respect to $\vlab$ yields
	\begin{equation}
		[\gradient_{\pred}(\vlab,\vpred)]_{i} = \sum_{k} [H_2(\vpred)]_{ik} [\vect{h}_1(\vlab)]_k + [\vect{h}_3(\vpred)]_{i} = [H_2(\vpred) \vect{h}_1(\vlab) + \vect{h}_3(\vpred)]_i \: ,
		\label{eq_intermediate}
	\end{equation}
	for some vector functions $\vect{h}_1: \labelspace \rightarrow \mathbb{R}\added{^d}$ and $\vect{h}_3: \labelspace \rightarrow \mathbb{R}\added{^d}$, with
	\[
		\nabla_{\lab_j}[\vect{h}_1(\vlab)]_k = [H_1(\vlab)]_{jk} \: .
	\]
	Since the minimum of $\loss(\vlab,\vpred)$ w.r.t.\ $\vpred$ is obtained for $\vpred=\vlab$, we need to have $\gradient_{\pred}(\vlab,\vlab) = \vect{0}$ for all $\vlab$, and thus
	\[
		\vect{h}_3(\vpred) = - H_2(\vpred) \vect{h}_1(\vpred) \: .
	\]
	Substitution into~(\ref{eq_intermediate}) gives
	\[
		\nabla_{\pred_i} \loss(\vlab,\vpred) = 	[\gradient_{\pred}(\vlab,\vpred)]_i = \sum_k [H_2(\vpred)]_{ik} \left([\vect{h}_1(\vlab)]_k - [\vect{h}_1(\vpred)]_k\right) = \left[H_2(\vpred) \left(\vect{h}_1(\vlab) - \vect{h}_1(\vpred)\right)\right]_i \: .
	\]
	
	Integrating out $\vpred$ by parts, directly incorporating that $\loss(\vpred,\vpred) = 0$ for all $\vpred$, yields
	\begin{eqnarray*}
		\loss(\vlab,\vpred) & = & \sum_k [\vect{h}_2(\vpred)]_k \left( 	[\vect{h}_1(\vlab)]_k - [\vect{h}_1(\vpred)]_k \right) + H_4(\vpred) - H_4(\vlab) \\
		& = & \vect{h}_2(\vpred)^\top \left( \vect{h}_1(\vlab) - \vect{h}_1(\vpred) \right) + H_4(\vpred) - H_4(\vlab) \: ,
	\end{eqnarray*}
	with vector function $\vect{h}_2: \labelspace \rightarrow \mathbb{R}^d$ satisfying
	\[
		\nabla_{\pred_i} [\vect{h}_2(\vpred)]_k = [H_2(\vpred)]_{ik} \: ,
	\]
	and scalar function $H_4: \labelspace \rightarrow \mathbb{R}$ then following from our choice of $h_1$ and $h_2$ through
	\begin{equation}
		\nabla_{\pred_i} H_4(\vpred) = \sum_k [H_1(\vpred)]_{ik} [\vect{h}_2(\vpred)]_k \: .
		\label{eq_constraint}
	\end{equation}
	To arrive at the form~(\ref{eq_generalized}), we choose $\vect{h}_1(\vpred) = \vectg(\vpred)$, $H_4(\vpred) = \added{-} A(\vectg(\vpred))$ and $h_2(\vpred) = -\nabla_{\vect{g}(\vpred)} A(\vectg(\vpred))$ for some scalar function $A: \vectg\added{(}{\labelspace}\added{)} \rightarrow \mathbb{R}$, and see that these choices indeed satisfy~(\ref{eq_constraint}).
	
	Fixing $\vlab$, the solution for $\vpred$ such that $\loss(\vlab,\vpred) = 0$ follows from $\vectg(\vpred) = \vectg(\vlab)$. Our requirement that the solution $\vpred=\vlab$ is unique for any choice of $\vlab$ implies that the mapping $\vect{g}$ is invertible. The strict convexity of $A$ then follows from the requirement that $\loss(\vlab,\vpred) \geq 0$.
	
	We only enforced the requirement that the loss function $\loss$ has a clean bias-variance decomposition for the special case of a single fixed label $\vlab$. This requirement already suffices to nail down $\loss$ to the class of $g$-Bregman divergences. The clean bias-variance decomposition for a distribution of labels $\vrlab$ we then get `for free', since it applies to any $g$-Bregman divergence: it does not imply any further restrictions.}

\writetheorem{bregman}

The proof can again be found in the appendix. Integrating the mixed derivative $\hessian_{\pred,\lab}$ from Lemma~\ref{th_factorization} twice, first with respect to $\vlab$ and then with respect to $\vpred$, we return to the original loss function $\loss$. The requirement of the identity of indiscernibles imposes constraints on the various components of this loss function, which are only satisfied by $g$-Bregman divergences.\deleted{ Because of the Lebesgue differentiation theorem, we can ignore potential singularities within the set $S$ of Lebesgue measure zero.}

\begin{corollary}
	Within the class of continuous, nonnegative loss functions \deleted{$\loss \in C^{2,1}(\labelspace \times \labelspace \setminus S)$} with the identity of indiscernibles, the `$g$-generalized' squared \deleted{Mahalanobis distance}\added{Euclidean distance}
	\[
	\loss(\vlab,\vpred) = \deleted{[\vectg(\vlab)-\vectg(\vpred)]^\top K [\vectg(\vlab)-\vectg(\vpred)]}\added{\|\vectg(\vlab)-\vectg(\vpred)\|^2} \: ,
	\]
	\deleted{where $K \in \mathbb{R}^{d \times d}$ is a positive definite matrix and}\added{with} $g: \labelspace \rightarrow \vectg(\labelspace)$ an invertible mapping, is the only symmetric loss function that has a clean bias-variance decomposition.
\end{corollary}
The proof follows directly from Theorem~\ref{th_bregman} and the fact that the squared Mahalanobis distance is the only symmetric Bregman divergence~\citep{boissonnat2010bregman}. \added{The squared Mahalanobis distance can be written as a `$g$-generalized' squared Euclidean distance via an invertible linear mapping, and since the composition of invertible mappings is again invertible, this yields the stated form.}

Since the squared \deleted{Mahalanobis}\added{Euclidean} distance does not satisfy the triangle inequality, this also implies that no metric can \deleted{have}\added{admit} a clean bias-variance decomposition. \added{In particular, well-studied losses such as the $0$-$1$ loss or the $L_1$ (Manhattan) distance cannot be decomposed in this way: there exists no central prediction that separates the bias term from a variance term completely independent of the labels. This provides a rigorous explanation of why previous attempts, such as~\citep{wolpert1997bias,domingos2000unified,james2003variance}, could not yield a decomposition with a clean separation of bias and variance.}

\begin{corollary}
	Consider the class of continuous, nonnegative loss functions \deleted{$\loss \in C^{2,1}(\labelspace \times \labelspace \setminus S)$ }with the identity of indiscernibles. If a loss function within this class has a clean bias-variance decomposition, then the central prediction (assuming no equality constraints) corresponds to an $\vectf$-mean prediction, that is,
	\[
		\vpredopt = \argmin_{\vlab \in \labelspace} \expectation_{\vrpred} 	\loss(\vlab,\vrpred) = \vectf^{-1} \left( \expectation_{\vrpred} \vectf(\vrpred) \right) \: ,
	\]
	for some invertible mapping $\vectf: \labelspace \rightarrow \vectf(\labelspace)$. And vice versa, if the central prediction derived from a loss function within the class boils down to an $\vectf$-mean prediction, then this loss function must have a clean bias-variance decomposition.
\end{corollary}
The proof from loss function to moment averaging follows because such a loss function must be a $g$-Bregman divergence, which has this property of moment averaging as shown in the proof of Theorem~\ref{th_gbregmanineq}. The proof from moment averaging to loss functions follows from~\citet{banerjee2005optimality}, who showed that Bregman divergences are the only loss functions for which (in our terms) the central label $\vlabopt$ is always equal to the average label $\expectation_{\vrlab} \vrlab$.

Our uniqueness result in Theorem~\ref{th_bregman} can be viewed as a generalization of earlier work by~\citet{hansen2000general}. They already argued that loss functions with a clean bias-variance decomposition must be of the form~(\ref{eq_singlelabel}). However, the analysis was restricted to loss functions that measure the error between univariate labels and predictions. Furthermore, although the analysis did start from distributions in the exponential family and followed the reasoning leading to~(\ref{eq_almost}), it did not establish the connection to Bregman divergences.

\section{Relaxing the Assumptions}\label{sec_relax}

We imposed some restrictions on the loss functions under consideration, most notably the identity of indiscernibles. In this section, we discuss whether these are necessary and how the bias-variance decomposition changes, and can to some extent be saved, when we relax them.

\deleted{We assumed the loss function $\loss$ to be continuous and sufficiently smooth. This assumption allowed us to take the second derivative $\hessian_{\lab,\pred}$ with respect to the label $\vlab$ and the prediction $\vpred$ without concern for its existence, and to apply Schwarz's theorem, ensuring the symmetry of the mixed partial derivatives. We permitted some leniency by not requiring the derivatives to exist across the entire domain, acknowledging that several popular loss functions exhibit such singularities. As it turns out, if these non-smooth loss functions have a clean bias-variance decomposition, they must also be $g$-Bregman divergences.

Technically, we do not exclude the possibility that there are even less smooth functions -- for example those not twice continuously differentiable across a domain with finite Lebesgue measure -- that still have clean bias-variance decompositions while not being a $g$-Bregman divergence. Further research might reveal whether such artifacts exist or, perhaps more in line with our current findings, can be definitively ruled out with a proof requiring fewer restrictions.}

The identity of indiscernibles, $\loss(\vlab,\vpred) = 0$ if and only if $\vlab = \vpred$, is quite heavily used throughout this paper. It can be relaxed to $\loss(\vlab,\vpred) = 0$ if and only if $\vlab = \vect{c}(\vpred)$, where $\vect{c}: \labelspace \rightarrow \labelspace'$ is an involution: a function such that $\vect{c} \circ \vect{c}$ is the identity. In other words, $\vect{c}(\vect{c}(\vlab)) = \vlab$ or equivalently $\vect{c}^{-1}(\vlab) = \vect{c}(\vlab)$ for all $\vlab$. Obvious examples of such functions include $\vect{c}(\vlab) = -\vlab$ and $\vect{c}((\lab_1,\lab_2)) = (\lab_2,\lab_1)$. If $\vect{c}$ is an invertible function but not an involution, we can transform the loss function into one with the identity of indiscernibles through a change of variables in either $\vlab$ or $\vpred$.

Related to the identity of indiscernibles is our assumption that labels $\vlab$ and predictions $\vpred$ have the same domain and hence the same dimensionality. We can relax this assumption, which is particularly relevant for loss functions derived from the log likelihood when the dimensionality of the observations is lower than that of the predictions.

\begin{boxedexample}{}
	A prototypical example of a loss function with unequal domains for the labels and the predictions is minus the log likelihood of the form~(\ref{eq_loglikelihood}) for a Gaussian distribution with canonical parameters $\tilde{\vpred} = (m/\variance, -1/(2\variance))$ and sufficient statistics $\vlab = \vphi(z) = (z,z^2)$. This loss function is popular for fitting so-called mean-variance estimation neural networks~\citep{nix1994meanvariance,sluijterman2024meanvariance} that output predictions for both the mean $m$ and the variance $\variance$, while receiving observation $z$.

	Following the line of reasoning that led to~(\ref{eq_almost}), we can rework minus the log likelihood, up to irrelevant constants independent of $\tilde{\vpred}$ and $\vpred$, to the form
	\begin{eqnarray}
		- \log p_{B}(z;\tilde{\vpred}) & = & - \vlab^\top \tilde{\vpred} + B(\tilde{\vpred})  \added{+ \; \text{const}} = A(\vlab) - \vlab^\top \vectf(\vpred) + B(\vectf(\vpred)) \added{+ \; \text{const}} = \nonumber \\
		& = & \bregman{A}(\vlab,\vpred) \added{+ \; \text{const}} \: ,
	\label{eq_infinite}
	\end{eqnarray}
	with appropriate choices for $A$ and $\vectf$, while for the moment ignoring the constraint that $t_2 = t_1^2$, that is, for arbitrary $\vlab \in \labelspace \setminus \labelspace_{\text{eq}}$. We can then define the central prediction through
	\[
		\vpredopt = \argmin_{\vlab \in \labelspace \setminus \labelspace_{\text{eq}}} \expectation_{\vrpred} \bregman{A}(\vlab,\vrpred) = \vectf^{-1} \left( \expectation_{\vrpred} \vectf(\vrpred) \right) \: ,
	\]
	yielding the usual
	\[
		\text{variance} = \min_{\vlab \in \labelspace \setminus \labelspace_{\text{eq}}} \expectation_{\vrpred} \bregman{A}(\vlab,\vrpred) = B(\vectf(\vpredopt)) - \expectation_{\vrpred} B(\vectf(\vrpred)) \: .
	\]

	When the constraint $t_2 = t_1^2$ is not ignored, the construction in~(\ref{eq_infinite}) fails because $A(\vlab)$ goes to infinity on the manifold $\vlab = (z,z^2)$. Luckily, not adding $A(\vlab)$ in~(\ref{eq_infinite}) still gives a perfectly valid bias-variance decomposition for a single observation $z$:
	\begin{equation}
		\underbrace{- \expectation_{\tilde{\vrpred}} \log p_{B}(z;\tilde{\vrpred})}_{\text{expected loss}} = \underbrace{- \log p_{B}(z;\opt{\tilde{\vpred}})}_{\text{bias}} + \underbrace{B(\opt{\tilde{\vpred}}) - \expectation_{\tilde{\vrpred}} B(\tilde{\vrpred})}_{\text{variance}} \: .
	\label{eq_expbiasvar}
	\end{equation}
	The first term on the right-hand side can be interpreted as the bias and the combination of the second and third term as the variance.

	This bias-variance decomposition, which applies to any log likelihood derived from a probability distribution within the exponential family, is completely in line with the bias-variance decomposition for likelihood-based estimators \citep{heskes1998bias}. Viewed as a loss function, $-\log p_{B}(z;\tilde{\vpred})$ is no longer guaranteed to be nonnegative. It is also not easy to see how to properly define the central observation $\opt{z}$ from a distribution of observations, nor how to decompose the expectation $\expectation_{Z} - \log p_{B}(Z;\opt{\tilde{\vpred}})$ into an intrinsic error and a bias.
	
	\added{The decomposition of \citet{heskes1998bias} in principle applies to arbitrary sets of distributions. The central prediction in that decomposition is the geometric average of the predictive distributions. For exponential families, this geometric average is again a member of the same family and the Kullback-Leibler divergence corresponds to a Bregman divergence, where the parameters of the exponential family play the role of labels and predictions (see Section~\ref{sec_bregman}). For distributions outside the exponential family, the Kullback-Leibler divergence itself still defines a Bregman divergence, but now expressed in terms of the probabilities themselves rather than a lower-dimensional parameterization; in this case, the decomposition still holds with the probabilities as labels and predictions.}
\end{boxedexample}

\added{Many classification losses similarly involve predictions and labels from different domains. Margin losses, which depend on the product of the label $\lab \in \{-1,1\}$ and the prediction score $\pred \in \mathbb{R}$ written as $\ell(\lab \pred)$ for some $\ell: \mathbb{R} \rightarrow \mathbb{R}^+$, do not satisfy the identity of indiscernibles for the loss itself. Nevertheless, \citet{wood2022margin} show that a subclass -- the gradient-symmetric margin losses -- can be expressed as Bregman divergences, either via their conditional excess risk or by anchoring the second argument to a label-dependent constant, and then do admit a clean bias-variance decomposition similar to~(\ref{eq_expbiasvar}). This connection to Bregman divergences is crucial: it explains why these losses admit a clean bias-variance decomposition, and suggests that without such a connection, no margin loss would have this property. In this sense, the results of \cite{wood2022margin} are perfectly in line with ours, providing a complementary perspective on how Bregman-type structure enables decomposability.}

\section{Conclusion}

Our analysis reveals that loss functions permitting a clean bias-variance decomposition \deleted{must take the}\added{can be expressed in the simplified} form
\begin{equation}
	\loss(\vlab,\vpred) = \vect{h}_1(\vpred)^\top \vect{h}_2(\vlab) + H_3(\vpred) + H_4(\vlab)
\label{eq_abstract}
\end{equation}
for certain choices of $\vect{h}_{1,2}: \labelspace \rightarrow \mathbb{R}^d$ and $H_{3,4}: \labelspace \rightarrow \mathbb{R}$. \added{This compact representation summarizes the same structure that appears, in more elaborate form, in the proof of our main Theorem~\ref{th_bregman}, and in the more specific forms given in~(\ref{eq_generalized}) and (\ref{eq_concise}).}

The key characteristic is the factorization of the interaction term between labels $\vlab$ and predictions $\vpred$ into an inner product of two separate functions. By imposing the natural requirements of nonnegativity and the identity of indiscernibles, we show that such loss functions must necessarily take the form of $g$-Bregman divergences. \deleted{Under mild regularity conditions, }$g$-Bregman divergences are then the only loss functions with a clean bias-variance decomposition.

\deleted{Although we cannot entirely rule out the existence of more exotic loss functions that do not take the form~(\ref{eq_abstract}) but still permit a clean bias-variance decomposition, it is challenging to envision what such functions might look like. Certain constraints -- such as the requirement that the variance term has a unique minimizer for any given distribution of predictions -- are direct consequences of the premise that the loss function admits a clean bias-variance decomposition and do not rely on any additional regularity conditions.}

The utility of bias-variance decompositions for studying the generalization of machine learning models has already been called into question, for example by \citet{brown2024biasvariance}, \deleted{which}\added{who} highlighted that the central prediction may not always be realizable by any machine learning model within the class under consideration. For instance, with single decision stumps, the central prediction $\vpredopt$ for squared error is an average over stumps, which cannot itself be represented by a single stump. Our finding, that clean bias-variance decompositions only apply to (minor variations of) Bregman divergences, adds to these concerns. While Bregman divergences are widely used and include prominent examples like squared error and cross-entropy loss, they are inherently sensitive to outliers due to their convexity in the label space \citep{vemuri2011total,zhang2023robust}. This convexity (or quasi-convexity when considering $g$-Bregman divergences) appears to be a necessary property for a clean bias-variance decomposition, implying that such decompositions do not extend to more robust loss functions that are less affected by outliers.

Although much broader than just the squared Euclidean distance, the class of loss functions for which clean bias-variance decompositions can be obtained is more special than perhaps previously believed. Future work could consider the generalization of these results to bias-variance decompositions in function spaces \citep{frigyik2008functional,gruber2023uncertainty} or bias-variance-diversity decompositions for ensembles \citep{wood2023unified}.

\added{\acks{Special thanks to Rob Brekelmans for directing attention to the work on rho-tau divergences, a reviewer for excellent feedback on an earlier version\errata{, and Jesse Geneson for spotting typos and catching boundary irregularities in Theorems 8 and 9 and Lemma 11.}}}

\ifthenelse{\boolean{pinapp}}{
\newpage
%\appendix
\section*{Appendix}

\setboolean{inappendix}{true}

In this appendix, we provide the remaining proofs for the propositions, lemmas, and theorems in the main text.

\calltheorem{ordering}
\calltheorem{optimal}
\calltheorem{reverse}
\calltheorem{means}
\calltheorem{bvgbregman}
\calltheorem{gbregmanineq}
\calltheorem{bregconstr}
\calltheorem{factorization}
\calltheorem{bregman}
}{}

\vskip 0.2in
\bibliography{bregmanrefs}

\end{document}